\documentclass[draftclsnofoot,onecolumn]{IEEEtran}

\usepackage{nomencl}
\makenomenclature
\usepackage{cite}

\usepackage[comma,numbers,square,sort&compress]{natbib}
\usepackage{amsmath,amsbsy}
\usepackage{amssymb}
\usepackage{float}
\usepackage{amsthm}
\usepackage{graphicx}
\usepackage{epstopdf}
\usepackage{color}
\usepackage{soul}
\usepackage{verbatim}
\usepackage{tikz,times}
\usepackage{multirow}
\usepackage{gensymb}
\usepackage{lipsum} 
\usepackage{subfig}
\usepackage[ruled,vlined]{algorithm2e}
\usepackage{footnote}

\newtheorem{remark}{\textbf{Remark}}

\usepackage{stackengine}

\title{\LARGE \bf Automated Driving Systems Data Acquisition and Processing Platform}

\author{ 
Xin Xia, Zonglin Meng, Xu Han, Hanzhao Li, Takahiro Tsukiji, Runsheng Xu, Zhaoliang Zhang, Jiaqi Ma

\thanks{Jiaqi Ma, Xin Xia, Zonglin Meng, Xu Han, Hanzhao Li, Takahiro Tsukiji, Zhaoliang Zhang, and Runsheng Xu are with UCLA Mobility Lab, Department of Civil and Environmental Engineering
Los Angeles, California 90095, USA.

}
\thanks{Manuscript received XX, XX, 2021; revised XX, XX, XX.}}

\begin{document}

\maketitle

\begin{abstract}
In this paper, an automated driving system (ADS) data acquisition and processing platform for vehicle trajectory extraction, reconstruction, and evaluation based on connected automated vehicle (CAV) cooperative perception are presented. This platform presents a holistic pipeline from the raw advanced sensory data collection to data processing, which is capable to process the sensor data from multi-CAVs and extract the objects’ Identity (ID) number, position, speed, and orientation information in the map and Frenet coordinates. First, the ADS data acquisition and analytics platform are presented. Specifically, the experimental CAVs platform and sensor configuration are shown and the processing software, including a deep-learning-based object detection algorithm using LiDAR information, a late fusion scheme to leverage cooperative perception to fuse the detected objects from multi-CAVs, and a multi-object tracking method is introduced. To further enhance the object detection and tracking results, high-definition maps consisting of point cloud and vector maps are generated and forwarded to a world model to filter out the objects off the road and extract the objects' coordinates in Frenet coordinates and the lane information. In addition, to refine trajectories from the object tracking algorithms, a post-processing method is proposed. Given the objects' information from object detection and tracking as well as the world model, a Kalman filter and Chi-square test method are applied to reduce the noise and remove the outlier in the trajectories. Aiming at tackling the ID switch issue of the object tracking algorithm, a fuzzy-logic-based approach is proposed to detect the discontinuous trajectories belonging to the same object. Then, a vehicle-kinematics-based trajectory prediction method is used, and a forward-backward-smoothing technique is applied to reconstruct the trajectory between the discontinuous trajectories. Finally, results, including object detection and tracking and a late fusion scheme, are presented, and the improvements by the post-processing algorithm in terms of noise level and outlier removal are discussed, which confirm the functionality and effectiveness of the proposed holistic data collection and processing platform. In another aspect, the extracted objects' information and generated HD maps can be used for several purposes in the transportation research community and ADS development community: analyzing the interaction between human-driven vehicles and ADS-equipped vehicles, car-following behavior analysis of ADS-equipped vehicles, traffic flow status analysis and modeling, and scenario generation for ADS testing. 
\end{abstract}


\IEEEpeerreviewmaketitle

\section{Introduction}\label{sec:introduction}

\begin{figure}[htb!]
\centering
\includegraphics[width=1\linewidth]{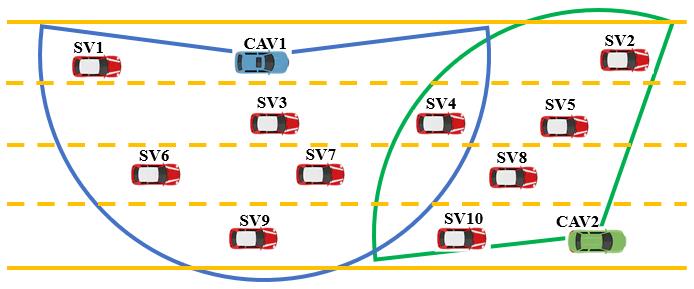}
\caption{Cooperative perception using 3D LiDAR. The blue and green vehicles are connected automated vehicle (CAV) 1 and CAV2, respectively. The surrounding vehicles (SV) are represented by the red vehicles. The areas within the blue and green sectors are covered by the LiDARs on CAV1 and CAV2, respectively. The yellow solid and dashed lines are the lane lines.}
\label{fig:Cooperative perception}
\end{figure}

With the new disruptive connected automated vehicles (CAVs) technology looming, researchers and engineers need to understand the benefits of this technology. In the meantime, they should prepare for the future challenges that these vehicles will encounter and potentially cause to traffic. Automated driving systems (ADS) vehicles equipped with multi-modal sensors (e.g., LiDAR, camera, RTK-corrected global navigation satellite system (GNSS)) for vision provide advanced sensor data to potentially perceive the traffic environments~\cite{xu2022v2x,xu2022cobevt}. An enormous amount of information can be extracted to understand new mixed traffic performance, such as ADS operational safety, the interaction between CAVs, and other traffic, for instance, traffic oscillations, heterogeneity, car-following behavior, and lane change behavior \cite{li2021operational,hu2022processing,han2022strategic}. Therefore, there is a need for an ADS data acquisition and processing platform to process the advanced sensor data in ADS-equipped vehicles or CAVs to obtain surrounding vehicles' trajectory data for ADS and transportation communities \cite{hu2022processing}.

In the literature, there have been extensive datasets for vehicle trajectory collection and vehicle interaction analysis. The most commonly used is the Next Generation Simulation program (NGSIM) dataset \cite{dot2016next}, which is collected by processing the video from cameras installed on infrastructure. Similar work for car-following behavior analysis in platooning scenarios was conducted by using aerial cameras or GNSS on each vehicle to obtain vehicle trajectories \cite{jiang2014traffic,wu2019tracking,zhaoopen}. Overall, these traditional datasets are mainly for analyzing human-driven vehicle behavior and traffic flow. To better understand the influence on traffic safety, reducing traffic congestion, and reducing energy consumption by involving individual ADS or CAV in traffic, efforts from both industry and academia have been made, and datasets such as KITTI \cite{geiger2013vision}, Waymo Open Dataset \cite{sun2020scalability}, Lyft Level 5 AV Dataset \cite{houston2020one}, nuScenes Dataset \cite{caesar2020nuscenes}, CADC Dataset \cite{pitropov2021canadian}, ApolloScape Dataset \cite{huang2018apolloscape} have been proposed. These datasets mainly include raw sensor data from advanced sensors such as LiDAR, camera, and GNSS/IMU (inertial measurement unit) integration system and the labeled data, including the bounding boxes in images and LiDAR point cloud. To further investigate the impact of CAVs on traffic, our team has published the first open datasets OPV2V \cite{xu2022opv2v} consisting of raw sensor data from LiDAR, cameras, and GNSS/IMU in multiple CAVs. These human-labeled datasets for both AVs and CAVs can be leveraged to extract the objects' (surrounding vehicles) trajectory. However, the AV and CAV technologies iterate fast and will co-exist with human-driven vehicles for a considerable amount of time \cite{sharma2021connected}, and different software stacks will have different influences on traffic. It is unrealistic to always use labor to label the datasets and analyze the influence whenever the software updates, as labeling the datasets is very expensive. On another aspect, researchers may have customized scenarios to collect data, and these existing datasets do not meet the diverse requirements. Despite these factors, the data format and structure of existing AV or CAV datasets are not easy to use as that of NGSIM because data for the different sensors are stored separately in a different format and in different coordinates \cite{hu2022processing}. It is hard to use the sensor data from AVs/CAVs to analyze their behavior of them. In this sense, there are need to have a platform that can collect the sensor data from these advanced sensors and process them to obtain the objects' trajectories that the transportation community needs.

Taking advantage of the development of CAVs' sensors, including 3D-LiDARs, cameras, radars, and GNSS are usually implemented. 3D-LiDAR has the capability to detect objects within a range at high accuracy. For example, as shown in Fig.~\ref{fig:Cooperative perception}, the surrounding vehicles within the blue sector will be detected by the 3D-LiDAR on the blue CAV. When there are multiple CAVs, the LiDAR data on different vehicles can be aggregated, and more surrounding vehicles will be detected, i.e., the sensor detection range of each vehicle can be extended, which is beneficial to construct the traffic flow on a larger scale. However, to the best of our knowledge, there is neither research that utilizes the 3D-LiDAR sensors on the equipped vehicle nor works which leverages the advanced sensors on multiple-CAVs to extract the surrounding vehicles' trajectories. This paper is trying to fill these gaps and propose a holistic and systematic platform that is able to collect multi-modal sensors from multiple CAVs and to process them to extract, reconstruct, and evaluate the trajectories. This platform is user-friendly to the transportation community and even the ADS community. It can be deployed according to customized needs. On top of that, the experiments which are for collecting sensor data already cover naturalistic city road scenarios and highway and freeway scenarios, which can be directly used for many purposes, such as car-following behavior analysis.

To summarize, we proposed an ADS data processing framework for vehicle trajectory extraction, reconstruction, and evaluation. The main contributions are listed as follows:

\begin{itemize}

\item {We proposed the first full data acquisition and analysis platform, which is able to collect automated driving system advanced sensor data and process them to extract the objects’ identity (ID) number, position, speed, and orientation information in different coordinates.}
\item{The data processing pipeline is capable of leveraging the sensor data from multiple CAVs and the world model with HD maps can assist the object detection and tracking algorithm in filtering out the off-road objects and obtaining the downtrack and crosstrack information in Frenet coordinates. To the best of our knowledge, this is the first framework that processes the real-world sensor data from multiple CAVs and provides the objects' information in the Frenet coordinates.}
\item{To further improve the objects' information, such as trajectories, the Kalman filter and the Chi-square test are applied to reduce noise and reject outliers in the trajectories; the proposed trajectory discontinuity detection method can identify the discontinuous trajectories and the trajectories can be reconstructed by our forward-backward-prediction smoothing method. }
\end{itemize}

The remainder of this paper is organized as follows: Section.~\ref{sec:ADS data acquisition and analytics Platform} introduced the ADS data acquisition and analytics platform.  Section.~\ref{sec:Data collection hardware and processing software} briefly shows the data collection hardware configuration and data processing software. The details for the LiDAR-based object detection, late fusion for detected objects from CAVs, multi-object tracking, HD map generation method, and world model are presented in Section.~\ref{sec:Object detection and tracking framework}. Section.~\ref{sec:Post processing} focuses on post-processing the vehicles' information by trajectory denoising and outlier rejection, trajectory discontinuity detection, and trajectory reconstruction method
Section.~\ref{sec:Results and discussion} provides the results and evaluation of the data processing pipeline. Section.~\ref{sec:conclusions} concludes this paper.

.

\section{ADS data acquisition and analytics Platform}\label{sec:ADS data acquisition and analytics Platform}

\begin{figure*}[htb!]
\centering
\includegraphics[width=1\linewidth]{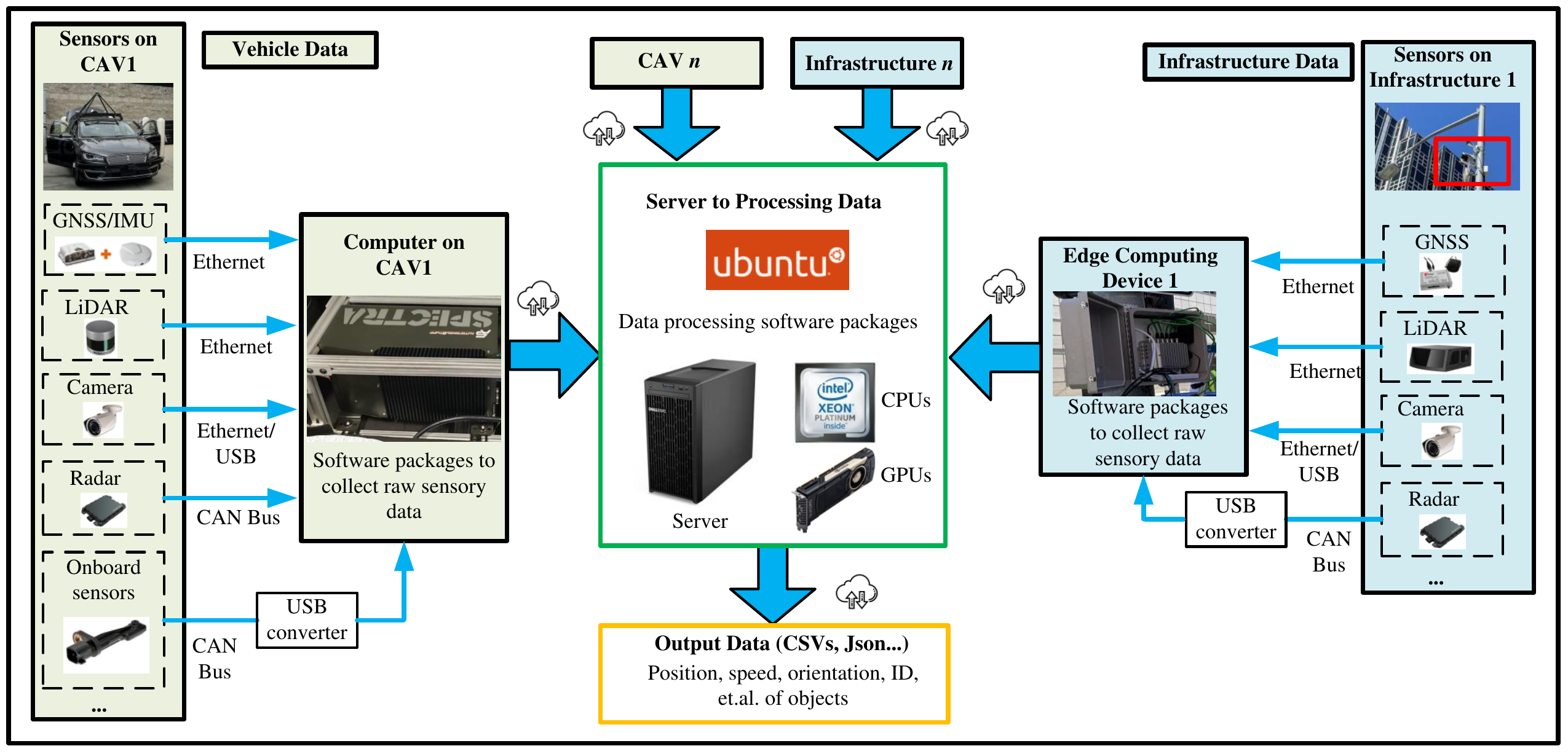}
\caption{ADS data acquisition and analytics platform. Sensor data from CAVs and infrastructures can be acquired by the computer on CAV or edge computing device at the infrastructure. Then, those sensor data will be forwarded to the server with multiple CPUs and GPUs and processed by the data processing software proposed in this paper. Data of objects' position, speed, orientation, and ID information will be output in the format of CSV or Json.}
\label{fig:ADS data acquisition and analytics platform}
\end{figure*}

Fig.~\ref{fig:ADS data acquisition and analytics platform} shows the ADS data acquisition and analytics platform proposed by UCLA Mobility Lab. Both the sensor data from CAVs and infrastructures can be collected and processed. CAVs are commonly equipped with multiple advanced sensors such as GNSS/IMU integration systems, LiDARs, Cameras, radars, and onboard sensors, including wheel speed sensors. GNSS/IMU integration system and LiDAR can be accessed through Ethernet ports. The camera can be connected by Ethernet or USB port of the computer. The radar and onboard sensors usually send data through the CAN (controller area network) bus. Then, a CAN bus to USB converter from Kvaser company can be applied to convert the CAN bus signal into a USB signal, which is readable by the computer. Those sensor data will be collected by the computer on the CAV with sensor drivers installed. As for infrastructure, GNSS, LiDARs, cameras, and radar are also equipped; the communication channel is the same as that for CAVs. The sensor data are collected by the edge computing device mounted at the infrastructure. Sensor data from CAVs and infrastructures will be processed by the software packages running in the Ubuntu system on the server, which has multiple CPUs and GPUs. Details about the software packages are presented in the following section of this paper. Then, the processed data, including position, speed, orientation, or ID information of the objects, will be output through CSVs or Jsons, which are user-friendly data formats and ease of usage.

Note that the proposed platform is a general platform and capable of processing data from individual/multiple AVs, CAVs, and infrastructures. The procedures to collect the data and the methodology to process sensor data from different agents are the same. In this paper, our real-world experiments focus on the sensor data from two CAVs. However, in the near future, infrastructure sensor data will also be collected and processed along with CAV sensor data.

\section{Data collection hardware and processing software}\label{sec:Data collection hardware and processing software}

\begin{figure}[htb!]
\centering
\includegraphics[width=1\linewidth]{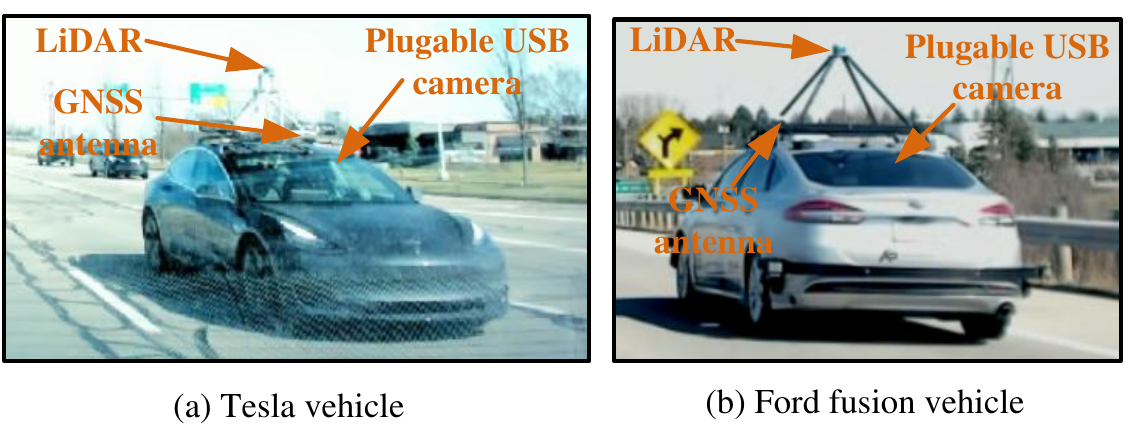}
\caption{Experimental CAVs. CAV1 is a retrofitted Tesla vehicle and CAV2 is a retrofitted Ford fusion.}
\label{fig: CAV figures}
\end{figure}

\subsection{Experimental vehicles configuration} \label{sec:Experimental vehicles configuration}

In this paper, two retrofitted experimental CAVs shown in Fig.~\ref{fig: CAV figures} are used to collect sensor data. A retrofitted Tesla vehicle  (CAV1) is shown in Fig.~\ref{fig: CAV figures}. a is equipped with extra sensors, including a 32-line Velodyne LiDAR (VLP-32c), two pluggable USB mono-cameras, and a GNSS /IMU integration system RT3000 product from OXTS company with RTK correction. One of the cameras is to collect the front view and the other is to collect the rearview. In addition to the Tesla vehicle, a retrofitted Ford Fusion vehicle (CAV2) from AutonomouStuff(AStuff) Company shown in Fig.~\ref{fig: CAV figures}.b is equipped with a 32-line Velodyne LiDAR (VLP-32c), two pluggable USB mono-cameras, and a GNSS/IMU integration system (Novatel SPAN product from Novatel Company). The cameras are configured the same as those on the Tesla vehicle. For example, the Tesla vehicle in Fig.~\ref{fig: CAV figures}. a is captured by the rear camera of the Ford Fusion vehicle and the Ford Fusion vehicle in Fig.~\ref{fig: CAV figures}.b is shot by the front camera of the Tesla vehicle. Each vehicle is equipped with a computer in the trunk to collect sensor data individually. When conducting the experiments, the two vehicles were simultaneously deployed as CAVs on the road. After the experiments, sensor data from different vehicles will be organized and synchronized to be processed by the proposed data processing pipeline introduced in Section.~\ref{sec:Object detection and tracking framework}.

\begin{remark}
It can be seen that the sensors, such as pluggable USB cameras and LiDAR, can be mounted to a production vehicle depending on the needs of the datasets without much effort. Another notice is that, although only two CAVs are deployed due to resource constraints, more vehicles can be deployed to collect data in parallel as the data processing pipeline can handle the data from more CAVs instead of just two.
\end{remark}

\subsection{Data processing software} \label{sec:Data processing software}

\begin{figure*}[htb!]
\centering
\includegraphics[width=1\linewidth]{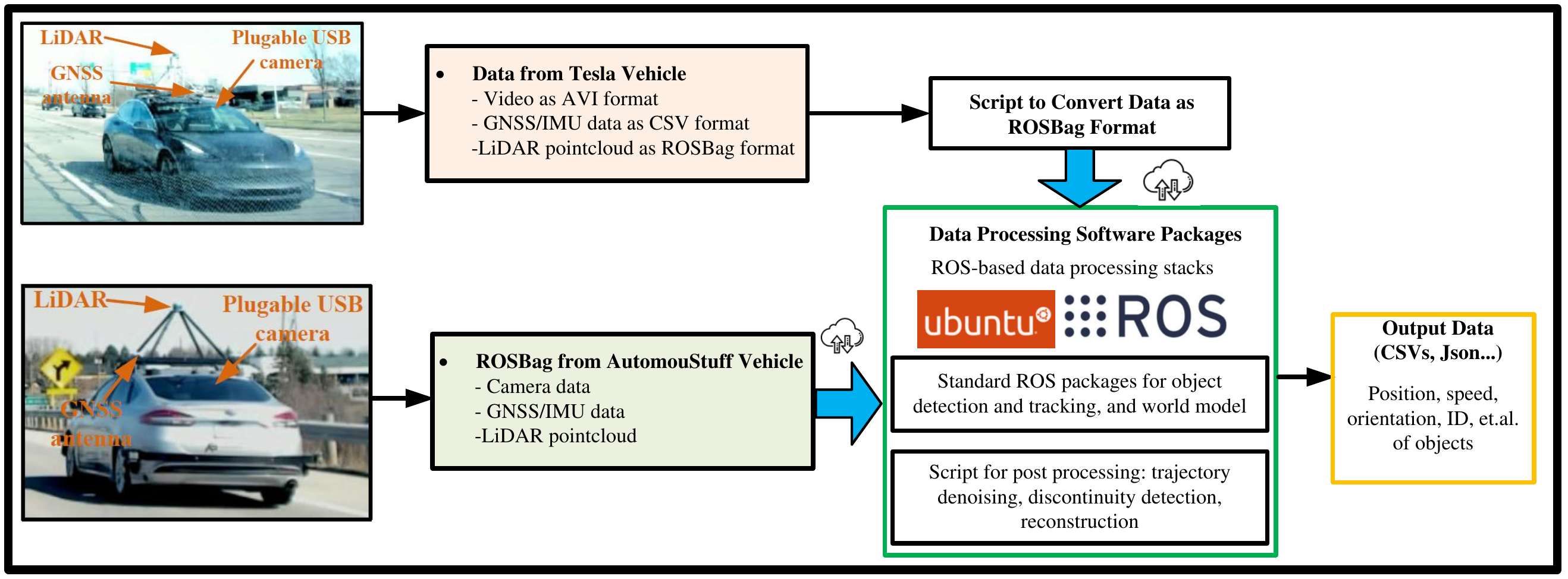}
\caption{Data flow of data processing software.}
\label{fig:Data flow of data processing software}
\end{figure*}

Fig.~\ref{fig:Data flow of data processing software} shows the data flow from the data collection by the two CAVs to the output. For the Tesla vehicle, the data from GNSS/IMU integration system are saved as CSV files, the point cloud is saved in ROSBag files, and the images from cameras are saved as AVI video files. All the sensor data of LiDAR, GNSS/IMU integration system, and camera from the AStuff vehicle are saved in ROSBag. To unify the source data from both vehicles, python scripts are used to convert the different data files from the Tesla vehicle into ROSBags. ROSBag is a standard data format that is used in Robot Operating System (ROS) \cite{quigley2009ros}. Then, the ROSBags from both vehicles will be input into the data processing software packages developed in ROS to extract the objects' information such as position, speed, orientation, and ID. These pieces of information will be output as files in the format of CSV or Json to make easy usage for scholars from different communities.

\begin{figure*}[htb!]
\centering
\includegraphics[width=0.7\linewidth]{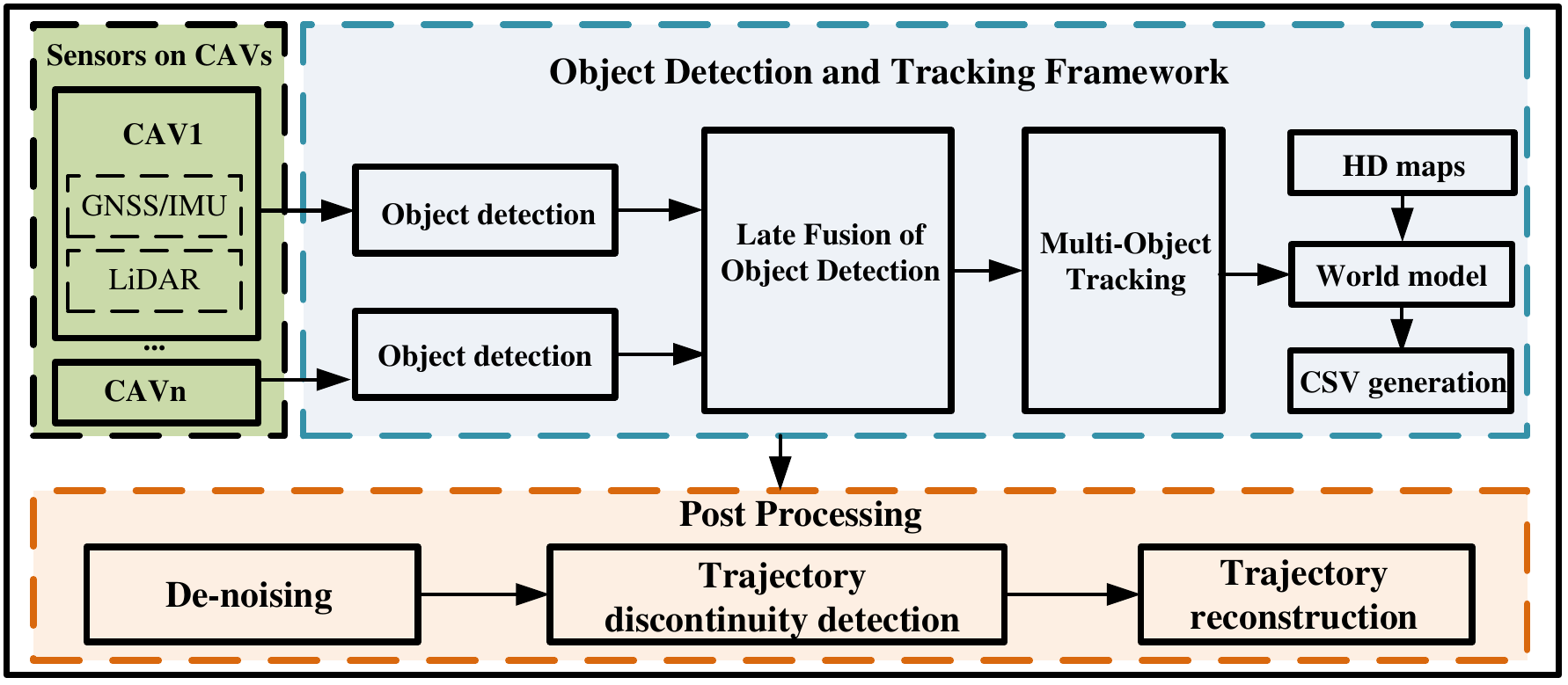}
\caption{Data processing pipeline. Sensory data from $n$ connected automated vehicles are processed to extract the trajectory, speed, orientation, and lane information of the objects, and that information will be saved in CSVs. Then, post-processing will be conducted on these CSVs to reduce the noise in the trajectory, speed, and orientation information, detect the discontinuity of the trajectory and reconstruct the trajectory. }
\label{fig: data processing pipeline}
\end{figure*}

\begin{figure*}[htb!]
\centering
\includegraphics[width=1\linewidth]{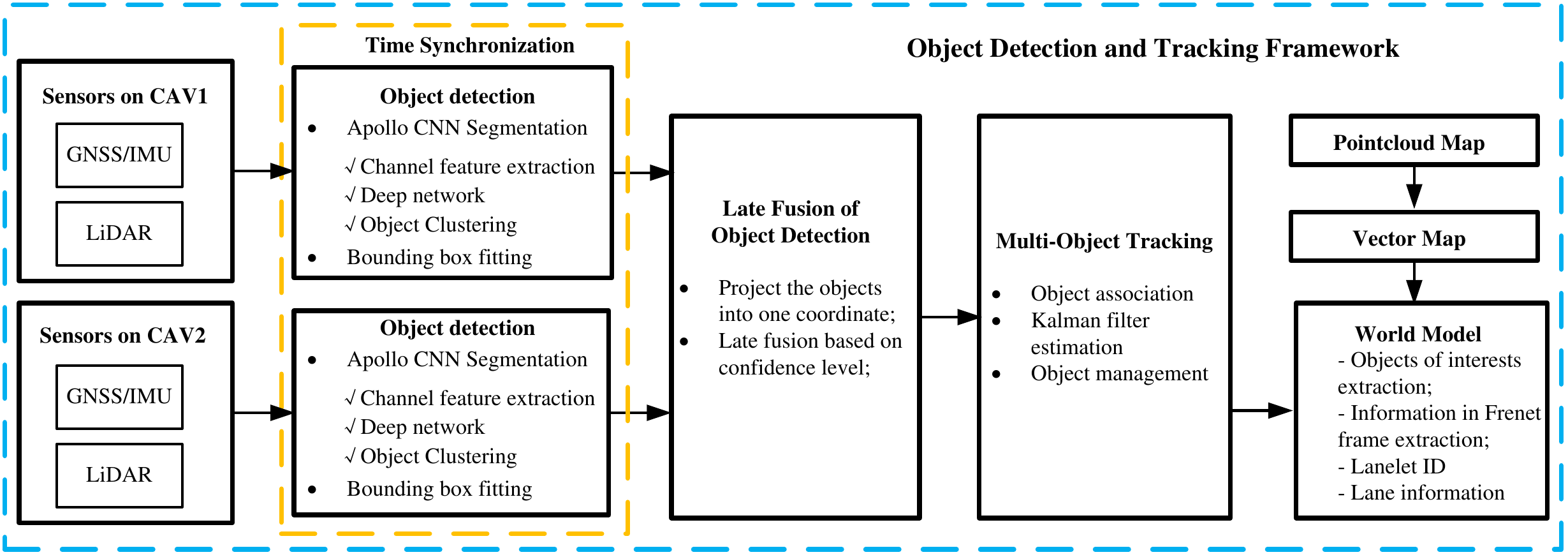}
\caption{Object detection and tracking framework. }
\label{fig: Object detection and tracking framework}
\end{figure*}

Details of the data processing software will be introduced in the following sections of this paper. In Fig.~\ref{fig: data processing pipeline}, the ADS data processing framework is shown. A 32-line 3D LiDAR and an RTK-corrected GNSS/IMU unit are implemented on the CAVs. With the point cloud from the 3D LiDAR, object detection will be performed on each CAV to generate the detection results (3D bounding boxes). Then, the detection results from both CAVs, which are time synchronized, are projected into one common coordinate, and the 3D bounding boxes from the two vehicles are merged through a late fusion method. With the fused detection results, 3D bounding boxes are tracked through a multi-object tracking algorithm, which will provide the IDs, position, speed, and orientation information of the corresponding 3D bounding boxes. Having the object tracking results, a Lanelet vector map which includes the lane information, will be used to extract the objects on the road. In the meantime, the coordinates (trajectory) of objects in the map and Frenet coordinate, which are defined in Appendix.~\ref{sec:map coordinate definition} and \ref{sec:frenet coordinate definition} and the lane information where the objects locate are computed by the world model. Then, the trajectory, speed, orientation, and lane information will be provided in the generated CSVs as datasets. In addition, in order to improve the quality of the datasets, post-processing is conducted by reducing the noise in the trajectory data, detecting the discontinuity in the trajectory data for those trajectories belonging to the same objects, and then reconstructing the trajectory.

Note that although in our datasets, we have two CAVs to collect the sensor data, the framework presented in Fig.~\ref{fig: data processing pipeline} can handle more than two CAVs to run simultaneously as the detected objects from an individual vehicle will be fused in a late fusion manner. Thus, the framework in Fig.~\ref{fig: data processing pipeline} is a general framework that works for processing the sensory data from $n$ ($n \ge 1$) CAVs. It is worth mentioning that in this framework, the object detection algorithm can be any state-of-the-art object detection algorithm such as PointPillar as long as the bounding boxes can be output. Similarly, the multi-object tracking algorithm can also be another object tracking algorithm such as \cite{luo2021exploring}.

\section{Object detection and tracking framework}\label{sec:Object detection and tracking framework}

An instantiation of the object detection and tracking framework in Fig.~\ref{fig: data processing pipeline} is proposed in Fig~\ref{fig: Object detection and tracking framework}. Two retrofitted CAVs are deployed to collect the sensory data, including 3D point cloud and GNSS/IMU data. The Apollo CNN segmentation algorithm is adopted to segment the point cloud from the individual CAV into clusters belonging to different objects. Then, 3D bounding boxes will be fitted based on the convex hull of the point cloud \cite{naujoks2018orientation}. Then, the bounding boxes detected by the object detection algorithm on the two CAVs will be fused in a late fusion framework. In addition, the fused 3D bounding boxes are tracked by a multi-object tracking algorithm to provide the objects' ID and estimate their position, speed, and orientation information. HD maps using the pipeline in Fig~\ref{fig:HD map generation pipeline} are generated and used to provide the coordinates of the detected objects in Frenet coordinates and the lane information in the world model.

\subsection{Object detection} \label{sec:object detection}

\subsubsection{Object detection based on Apollo CNN segmentation}

\begin{figure}[htb!]
\centering
\includegraphics[width=1\linewidth]{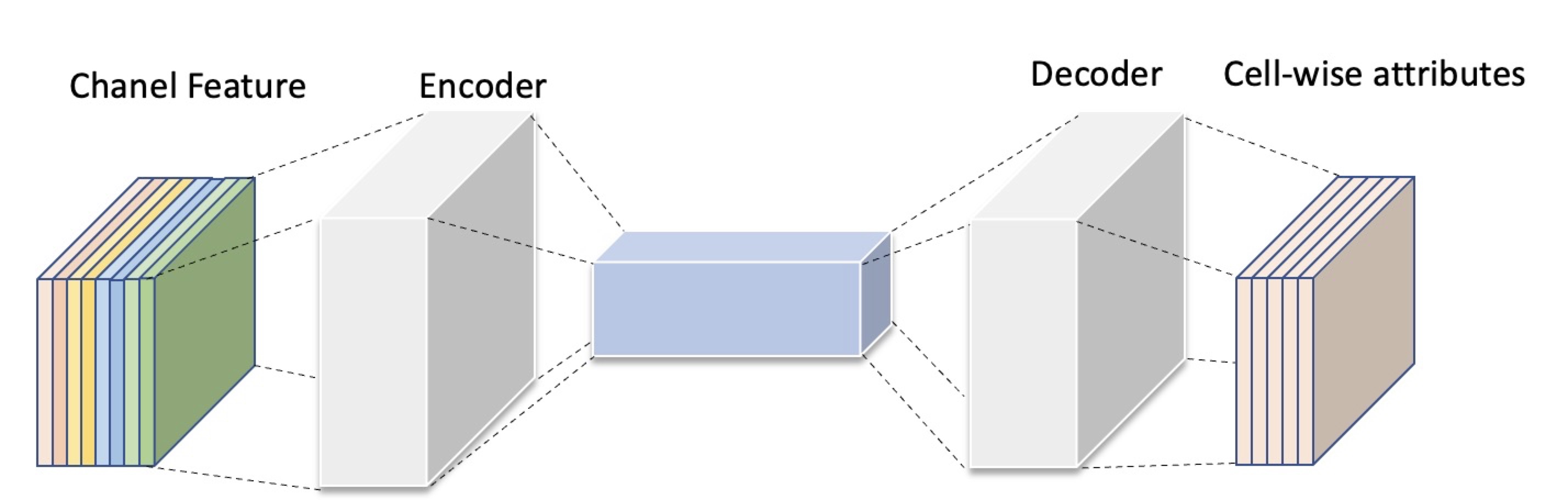}
\caption{Baidu CNN Segmentation.}
\label{fig: Baidu CNN Segmentation}
\end{figure}

We propose to leverage the tracking-by-detection perspective in this paper. The tracking-by-detection framework is essentially made of an object detector and a data association algorithm that establishes track-to-detection correspondence.

For object detection, the Apollo CNN segmentation \cite{apollo} to segment out the vehicles in each point cloud frame is employed. The task of segmentation is to classify the point clouds into separate clusters. Apollo CNN segmentation accepts point cloud as input and segments out the objects. The segmentation consists of three main stages: (1) Channel Feature Extraction, which divides the Bird-Eye view projection of the point cloud into cells and calculates the static measurements of each cell; (2) Encoder-decoder Network that takes the measurements and predicts the necessary attributes for each cell; (3) A cell clustering algorithm that finds the candidate clusters and a post-processing method to filter out some candidates clusters.

\textbf{Channel Feature Extraction:}  To segment the point cloud objects and apply a 2D convolutions architecture, Apollo CNN segmentation first converts the LiDAR point clouds into 2D BEV (birds eye view) grid space and represents the point cloud by statistical measurements. Specifically, each cell in grid space computes 8 different measurements of the point clouds, including the maximum height of points in the cell, intensity of the highest point in the cell, mean height of points in the cell, mean intensity of points in the cell, a number of points in the cell, angle of the cell’s center with respect to the origin, the distance between the cell’s center and the origin, and a binary value indicating whether the cell is empty or occupied. The whole measurements $A \in \mathbf{R}^ {W \times H \times 8}$ will be used as the network input, where the W and H represent the number of rows and columns of the grid and 8 is the statistical measurements for each cell.

\textbf{Deep Network :}
The network used by Apollo consists of an encoder and decoder as shown by Fig \ref{fig: Baidu CNN Segmentation}. The network will take the channel feature measurement $A$ as input and predict 5 cell-wise attributes including the center offset, objectness, positiveness, object height, and class probability. The center offset of each cell is essentially a vector point to the center of objects. The objectness indicates whether a cell contains objects. The positiveness and object height is used to filter out the background cluster and remove the point that is too high. The class probability will be used to classify each cluster. The encoder maps the computed channel feature $A$ to an abstract feature map, whereas the decoder takes the feature map as input, processes it, and produces the attributes $Y \in \mathbf{R}^ {W \times H \times 5}$.

\textbf{Object Clustering :}
After obtaining the network output attributes, Apollo clusters cells based on their center offset. Concretely, it first judges whether the current cell contains an object and then clusters the adjacent cells pointed by the current cell's center offset. After Apollo iterates all the cells, it will generate a number of candidate clusters which include several cells. The post-processing of Apollo could remove some candidate clusters having a small number of points or a low confidence score.

\begin{remark}
From our experience, when applying Apollo CNN segmentation, a few objects are missed. However, at some point clouds from small buildings, trees, or large garbage bins might be erroneously classified as vehicles. Developing a more accurate object detection model/algorithm or using other techniques is the proper approach to compensate for these falsely detected objects. In this work, we apply HD maps to filter out the missed classified off-road objects.

Another limitation of the LiDAR-based object detection algorithm is that the detection range is limited. For example, for the used 32-line Velodyne in our application, the detection range with high confidence is only 40-60m. Objects out of this range can not be detected stably. To address this issue, other sensors such as camera/radar or LiDAR with more lasers may be needed. In our future work, we will head to both of these two approaches to improve object detection performance.

\end{remark}

\subsubsection{Bounding box fitting}
After having the point cloud of the clusters from the Apollo segmentation algorithm, the bounding boxes will be generated by a 3D bounding box fitting algorithm in \cite{naujoks2018orientation}.

\begin{figure}[htb!]
\centering
\includegraphics[width=0.7\linewidth]{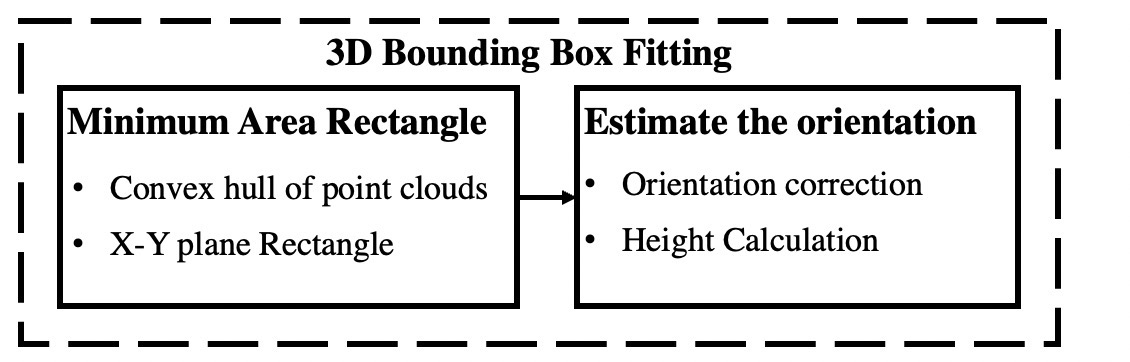}
\caption{Bounding box fitting method.}
\label{fig: box_fitting}
\end{figure}

Bounding box fitting could be defined as given a set of 3D segmented objects, we want to calculate oriented minimum area rectangles that contain the segmented point cloud. The algorithm mainly consists of constructing the minimum area rectangle and correcting the orientation of the rectangle as shown by Fig.~\ref{fig: box_fitting}. 
The method first projects the 3D point cloud into the 2D plane. After removing the interior points, the Graham-Scan algorithm is used to create the convex hull of the point clouds.

The Rotating Calipers algorithm is applied to construct the minimum area rectangle of the convex hull. The advantage of using convex hull representation is reducing the computation time. Although the generated rectangle covers the object with the minimum area, the orientation of such rectangles is not correct. Therefore, a three-point-based heuristic approach is used to calculate the orientation, which can be used to rotate the rectangles and correct the angle. The height of the bounding box is simply calculated from the difference between the maximum and minimum height of the clusters. 


\begin{remark}
Due to the sparsity of the point cloud, the clustered point cloud from a specific detected object might be only a small part of the object, leading to an error when fitting 3D bounding boxes. This error will cause the position and orientation errors of the corresponding object as the position of the object is assumed to be the center of the bounding box. These errors in position and orientation will be reduced by the multi-object tracking in this subsection and the trajectory de-noising algorithms in Section.~\ref{sec:trajectory denoising and ourlie rejection}.

In another aspect, to improve the performance of the bounding box, historical information from the LiDAR or semantic information from the camera may help because as long as the object ID does not change, the best predicted bounding box in the past frames can be found and represent the bounding box of the object instead of predicting the bounding box whenever the LiDAR point cloud is updated. 
\end{remark}

\subsection{Late fusion of object detection} \label{sec:Late fusion of object detection}
\subsubsection{Project objects into one common coordinate}
The bounding boxes fitting will be performed in each CAV. To fuse the bounding boxes (detected objects) from the object detection algorithm in the two CAVs, the bounding boxes will be temporally synchronized according to the timestamp of the LiDAR frames and projected into a common coordinate. Note that there is no difference to select either the coordinate in CAV1 or CAV2 as the common coordinate. In our application, the LiDAR on CAV2 is chosen as the common coordinate. The projection from CAV1 to CAV2 can be computed through \eqref{eqn:projection method}

\begin{align}
    \setlength{\arraycolsep}{4pt}
    \boldsymbol{T}_\text{CAV1}^\text{CAV2}= {\boldsymbol{T}_\text{CAV2}^{m}}^{-1}  \boldsymbol{T}_\text{CAV1}^{m},
    \label{eqn:projection method}
\end{align}
where superscript $m$ denotes the map coordinate and subscript CAV1 and CAV2 denotes the CAV1 and CAV2 coordinate respectively. $\boldsymbol{T}_{CAV2}^{m}$ and $\boldsymbol{T}_{CAV1}^{m}$  are transformations from the CAV2 coordinate to the map coordinate and from the CAV1 coordinate to the map coordinate. $\boldsymbol{T}_{CAV1}^{CAV2}$ is the transformation from CAV1 to CAV2.

\begin{align}
    \setlength{\arraycolsep}{4pt}
    \boldsymbol{T}_\text{CAV2}^{m}={\left [\begin{array}{cc}
         \boldsymbol{R}_\text{CAV2}^{m} & \boldsymbol{t}_\text{CAV2}^{m}  \\
         \boldsymbol{0} & 1
    \end{array}
    \right]},
\end{align}
where the rotation matrix $\boldsymbol{R}$ and translation vector $\boldsymbol{t}$ are defined by \eqref{eqn:rotation matrix}
and \eqref{eqn:translation vector}, respectively. $\phi$, $\theta$, and $\varphi$ are the roll, pitch, and heading angles between the CAV2 LiDAR frame coordinate and the map coordinate, respectively. $\text{c} \ast = \cos (\ast)$ and $\text{s} \ast = \sin (\ast)$.

\begin{align}
    \setlength{\arraycolsep}{4pt}
    \boldsymbol{C}_\text{CAV2}^{m}={\left [\begin{array}{ccc}
         \text{c}\varphi \text{c}\theta & -\text{s}\varphi \text{c}\phi+\text{c}\varphi \text{c}\theta \text{s}\phi  & \text{s}\varphi \text{s}\phi + \text{c}\varphi \text{s}\theta \text{c}\phi  \\
         \text{s}\varphi \text{c}\theta & \text{c}\varphi \text{c}\phi + \text{s}\varphi \text{s}\theta \text{s}\phi & -\text{c}\phi \text{s}\phi + \text{s}\varphi \text{s}\theta \text{c}\phi \\
        -\text{s}\theta   & \text{c}\theta \text{s}\phi   & \text{c}\theta \text{c}\phi 
    \end{array} \right]}, 
    \label{eqn:rotation matrix}
\end{align} 

\begin{align}
    \setlength{\arraycolsep}{4pt}
    \boldsymbol{t}_\text{CAV2}^{m}=[t_x, t_y, t_z]^\top , 
    \label{eqn:translation vector}
\end{align}
whereas $t_x$, $t_y$ and $t_z$ are the translation in the $x$, $y$, and $z$ directions, respectively. The angles including yaw, pitch, and roll can be accessed from the GNSS/IMU integration system on the vehicle. The translation vector is obtained by converting the longitude, latitude, and altitude from the GNSS/IMU integration system into $x$, $y$, and $z$ in a UTM (Universal Transverse Mercator) coordinate, which is the map coordinate in our application.

\subsubsection{Late fusion strategy}\label{sec:late fusion strategy}

After projecting the bounding boxes into a common coordinate, they will be aggregated together. The IoU (intersection of unit) between each bounding box and others is computed. If the IoU is below a threshold, the bounding box is kept indicating an object is only detected by one CAV. When the two vehicles are near each other, there is an intersection area where the bounding boxes will be predicted by the object detection algorithms on both CAVs. In this intersection area, for the detected vehicles (such as SV4 in Fig.~\ref{fig:Cooperative perception}), when the IoU of the corresponding bounding box is higher than a threshold, the bounding box with a higher confidence level which is predicted in the object detection algorithm is kept for the corresponding bounding box.  

\begin{remark}
By fusing the objects detected by both CAV1 and CAV2, the sensing range of each CAV can be extended, making it possible to analyze interaction between more CAVs and SVs and to create complex scenarios for ADS/CAV development. 
It is worth mentioning that the projection matrix given by \eqref{eqn:projection method} implies the assumption that the LiDAR and GNSS/IMU systems are time synchronized both temporally and spatially. Therefore, in real applications, attention should be paid to ensuring the synchronization conditions are met.
 
\end{remark}

\subsection{Object tracking}

\begin{figure}[htb!]
\centering
\includegraphics[width=1\linewidth]{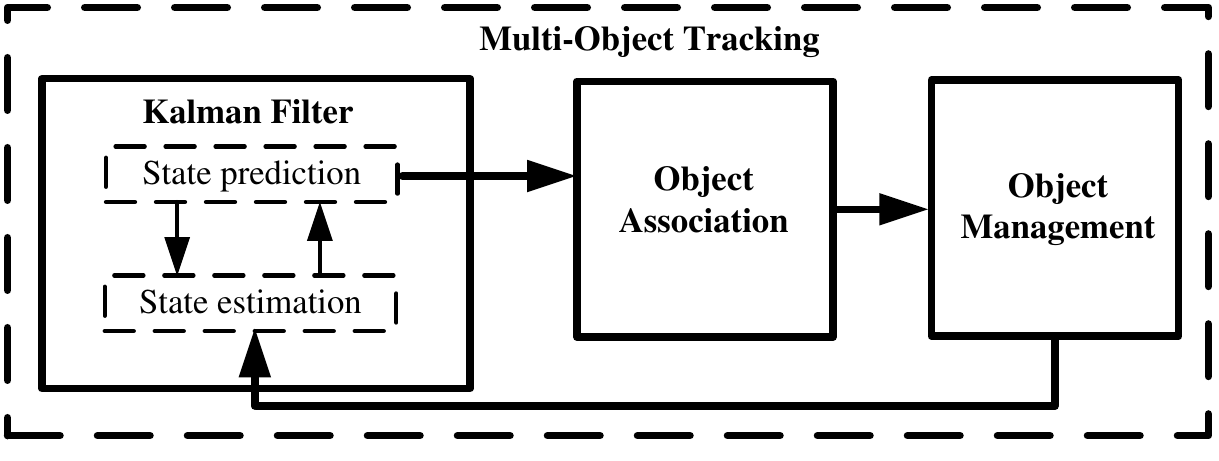}
\caption{Multi-object tracking algorithm.}
\label{fig:Multi-object tracking algorithm}
\end{figure}
Once the detected objects from CAVs are fused, the object tracking module will track them. The object tracking pipeline is shown in Fig.~\ref{fig:Multi-object tracking algorithm}. The bounding boxes from the late fusion algorithm will be input to an object tracking algorithm to track the objects, i.e., providing the estimation of the position, speed, and orientation of the corresponding object in the map coordinates and also its unique ID. Specifically, the input of the tracking algorithm is 3D bounding boxes $D = (x_v,y_v,z_v,l_v,h_v,w_v,\varphi_v)$ \cite{himmelsbach2012tracking}, where $x_v,y_v,z_v$ are the position of the bounding box, $l_v,h_v, \text{and}\ w_v$ are the length, height, and width of the bounding box, and $\varphi_v$ is the heading angle. The subscript $v$ means the variable in vehicle coordinate which is defined Section.~in \ref{sec:vehicle coordinate definition}. The objective of the tracking algorithm is to associate the detected 3D bounding boxes $D_t^i$ with tracks and obtain the object's states $\boldsymbol{x} = (x_m,y_m,z_m,l_v,h_v,w_v,\varphi_m,v_{x_m},v_{y_m},v_{z_m}, a_{x_m},a_{y_m},a_{z_m},\text{ID})$ for each vehicle, where $v_{x_m},v_{y_m},v_{z_m}$ are the velocity in 3D space, $a_{x_m}, a_{y_m}, a_{z_m}$ are the accelerations, ID is the unique ID number of each vehicle. The subscript $m$ denotes the variable in the map coordinate.

The object tracking algorithm mainly consists of three stages:(1) Kalman filter (State estimation), (2) Object association, (3) Object Management. The Kalman filter has two stages to estimate the states of the object. It will predict the state $\boldsymbol{x}(k+1)$ in the next step $k+1$ of an existing trajectory by the Kalman filter method.
\begin{align}
\boldsymbol{x}(k+1) = \text{KF}(\boldsymbol{x}(k))
\end{align} 
where KF means Kalman filter. After having the prediction of the states of each object, the Hungarian algorithm is used to associate the prediction with an existing object. Specifically, the Euclidean distance between the predicted position of the tracked objects and the current detection results is calculated and then, each detected object $D_{i} \in D_{t}$ can be associated with the nearest euclidean box  $D_{j} \in T_{t}$.

The object tracking management will add new objects and delete old objects. New objects are added if detected objects cannot be matched to any existing objects during the association process. Similarly, the object will be deleted if it is not associated with any detected objects after a period of time. A unique ID number will be assigned to the newly added objects and correspondingly, their states such as position, speed, and will be estimated.

In addition, the tracked objects will be input into the world model which will be introduced in Section.~\ref{sec:world model} to check if they are of interest, i.e. on the road or not. Then, the variables $(x_m,y_m,l_v,h_v,w_v,\varphi_m,v_m, a_m,\text{ID})$ including position, speed, acceleration, heading information and ID of each object on the road in the map coordinate, which is defined in Section.~\ref{sec:map coordinate definition}, are output into files with CSV format.

\begin{remark}
In a real application, the object might go beyond the LiDAR detection range, which is usually 40-60m, for a relatively long term and then, comes back within the detection range. In this process, the object will be removed from the object list by the object management module.  When it appears again, a new object will be added to the object list. This is a common ID switch issue of object tracking \cite{feng2019multi}. Although the objects are associated by the Hungarian algorithm, a long time of disappearing and appearing again will lead to association failure. To address this issue, after having the trajectories of all the objects, the trajectory discontinuity will be detected and the trajectories belonging to the same objects will be connected in the post-processing procedure, which is introduced in Section.\ref{sec:Trajectory discontinuity detection}. Another solution could be to leverage other sensors such as cameras to extract the semantic information to associate the objects to perform long-term object tracking \cite{tian2019online}.
\end{remark}

\subsection{HD map generation}
As stated in the object detection section, the objects which are off-road and of no interest might be detected erroneously. In this case, the HD maps can be used to refine the object results by filtering out the erroneously detected objects. Thus, this section will provide the HD map generation pipeline and accordingly show how HD maps are used in our data processing framework.

\begin{figure*}[htb!]
\centering
\includegraphics[width=1\linewidth]{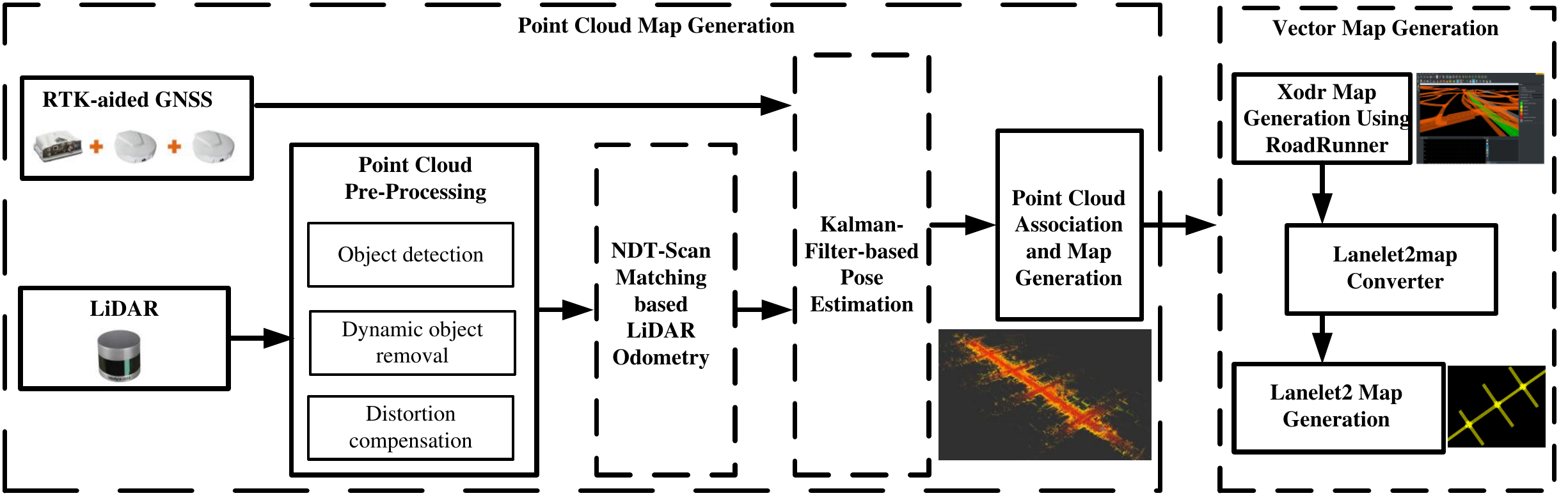}
\caption{HD map generation pipeline.}
\label{fig:HD map generation pipeline}
\end{figure*}

In Fig.~\ref{fig:HD map generation pipeline}, the  HD map generation pipeline including the generation of the point cloud and vector maps is shown. The point cloud from the LiDAR is pre-processed to remove the point cloud belonging to dynamic objects and compensate for the distortion in the point cloud. Then, an NDT (normal transformation distribution)-scan matching algorithm is applied to compute the relative transformation between two consecutive LiDAR frames. Then, LiDAR odometry is constructed to provide the pose of the LiDAR, and then, the pose is fused with the pose from an RTK-aided GNSS module within a Kalman filter. Taking the fused pose, the point clouds are transformed into the map coordinates and aggregated to generate the point cloud map. In addition, this point cloud map is imported in Roadrunner and used to provide lane information of the road to generate the Xodr map which will be converted to a vector map through a Xodr to Lanelet map converter.

\subsubsection{Point cloud map generation} When generating the point cloud map, the LiDAR frame will be pre-processed before it is used to generate point cloud map. The pre-processing of the LiDAR point cloud includes compensating for the distortion due to the movement of the subject vehicle and removing the dynamic vehicles on the roads. The object detection method in Section. \ref{sec:object detection} is used to detect the dynamic vehicles. Once the dynamic vehicles are detected, the point cloud on these vehicles will be removed from the current LiDAR frame. Then, combining the velocity and angular rate information, the distortion in the point cloud without dynamic vehicles will be compensated. Taking the pre-processed point cloud, an NDT (normal distribution transformation algorithm) scan matching algorithm is applied to associate the two consecutive LiDAR frames and compute the relative transformation between them. The current LiDAR points residing within a cell are transformed into a normal distribution. For cell $j$, the mean vector $\boldsymbol{q}$ of the points and the covariance matrix $\boldsymbol{C}$ are computed by \eqref{eqn:NDT_mean_vector} and \eqref{eqn:NDT_covariance}.

\begin{align}
    \setlength{\arraycolsep}{4pt}
    \boldsymbol{q}=\frac{1}{n}\sum_{j=1}^n\boldsymbol{x}_i ,
    \label{eqn:NDT_mean_vector}
\end{align}

\begin{align}
    \setlength{\arraycolsep}{4pt}
    \boldsymbol{C}=\frac{1}{n-1}\sum_{j=1}^n(\boldsymbol{x}_i-\boldsymbol{q})(\boldsymbol{x}_i-\boldsymbol{q})^\top ,
    \label{eqn:NDT_covariance}
\end{align}
where $\boldsymbol{x}_i$ represents the point $i$ residing in cell $j$, $i$ is the number of points in the cell $j$. Based on a certain transformation in \eqref{eqn:consecutive transformation}, for points in the current LiDAR frame, the coordinates of the transformed points in the previous LiDAR frame coordinates ${L'}$ are 
\begin{align}
    \boldsymbol{x}' = \boldsymbol{R}_{L'}^{L}\boldsymbol{x}+\boldsymbol{t}_{L'}^{L} ,
    \label{eqn:consecutive transformation}
\end{align}
 The NDT scan matching algorithm is to search for the best transformation $(\boldsymbol{C}_{L'}^{L}, \boldsymbol{t}_{L'}^{L})$ in the transformation function for the points in the current LiDAR frame to match the points in the previous LiDAR frame. The cost function for the parameters in the transformation function is defined as 
\begin{align}
    \setlength{\arraycolsep}{4pt}
    J(\boldsymbol{C}_{L'}^L,\boldsymbol{t}_{L'}^L)=-\sum_{j=1}^n exp \frac{(\boldsymbol{x}_j'-\boldsymbol{q}')^\top \boldsymbol{C}'^{-1}(\boldsymbol{x}_j'-\boldsymbol{q}')}{2} ,
    \label{eqn:NDT_score_function}
\end{align}
where $\boldsymbol{q}'$ and $\boldsymbol{C}'$ are the corresponding mean vector and covariance matrix of the point cloud in the previous LiDAR frame. Newton's algorithm is used to iteratively solve the optimization problem to find the best transformation between the current LiDAR coordinates and the previous LiDAR frame coordinates. Taking the transformation \begin{align}
    \setlength{\arraycolsep}{4pt}
    \boldsymbol{T}_{L'}^L={\left [\begin{array}{cc}
         \boldsymbol{C}_{L'}^L & \boldsymbol{t}_{L'}^L  \\
         \boldsymbol{0} & 1
    \end{array}
    \right]},
    \notag
\end{align}, the LiDAR odometry can be constructed as $\boldsymbol{T}_{L'(0)}^{L(k)}=\boldsymbol{T}_{L'(k-1)}^{L(k)}...\boldsymbol{T}_{L'(1)}^{L(2)}\boldsymbol{T}_{L'(0)}^{L(1)}$. However, due to the noise contained in the LiDAR frame, there are errors in $\boldsymbol{T}_{L'}^L$. These errors will be accumulated in the transformation $\boldsymbol{T}_{L'(0)}^{L(k)}$. To compensate for these errors, the translation and heading information provided in an RTK(real time kinematics)-aided GNSS module is used to provide the measurements for the transformation from the LiDAR odometry as the information from the GNSS module is free of accumulated errors. To fuse the pose from the LiDAR odometry and from the GNSS module, a Kalman filter is applied. Details of the Kalman filter can be found in \cite{xia2022estimation}. Then, given the fused pose from the Kalman filter, all the points in the LiDAR frames will be transformed into the map coordinates for generating the map. The transformation function is defined as 
\begin{align}
    \boldsymbol{T}(\boldsymbol{x}) = \boldsymbol{C}_L^m\boldsymbol{x}+\boldsymbol{t},
    \label{eqn:transformation}
\end{align}
where $\boldsymbol{x}=[x, y, z]^\top$ denotes the coordinates of the points in a LiDAR frame of the 3D LiDAR, $\boldsymbol{C}_L^n$ is the rotation matrix between LiDAR coordinates and map coordinates, and $\boldsymbol{t}=[t_x, t_y, t_z]^\top$ is the translation vector. Having the point cloud in the map frame, the LiDAR frames from time $t_0 \ \text{to} \ t_k$ are aggregated to the LiDAR point cloud map. Note that, in a real-time application, because the range of LiDAR sensor is large, there is no need to aggregate all the LiDAR frames to generate a map because involving all the point cloud frames in the map will be too large consuming much space of hard drive and computational resource. After the ego vehicle travels a certain distance (5m in our application), adding a LiDAR frame into the point cloud map is sufficient.

\begin{remark}
When generating the point cloud map, the point cloud belonging to dynamic objects needs to be removed because the point cloud should only contain static features. Another purpose is that the clean map without dynamic objects makes the lane lines more visible in terms of the intensity information of the point cloud when drawing vector maps in Roadrunner.
\end{remark}

\subsubsection{Vector map generation}

\begin{figure}[htb!]
\centering
\includegraphics[width=1\linewidth]{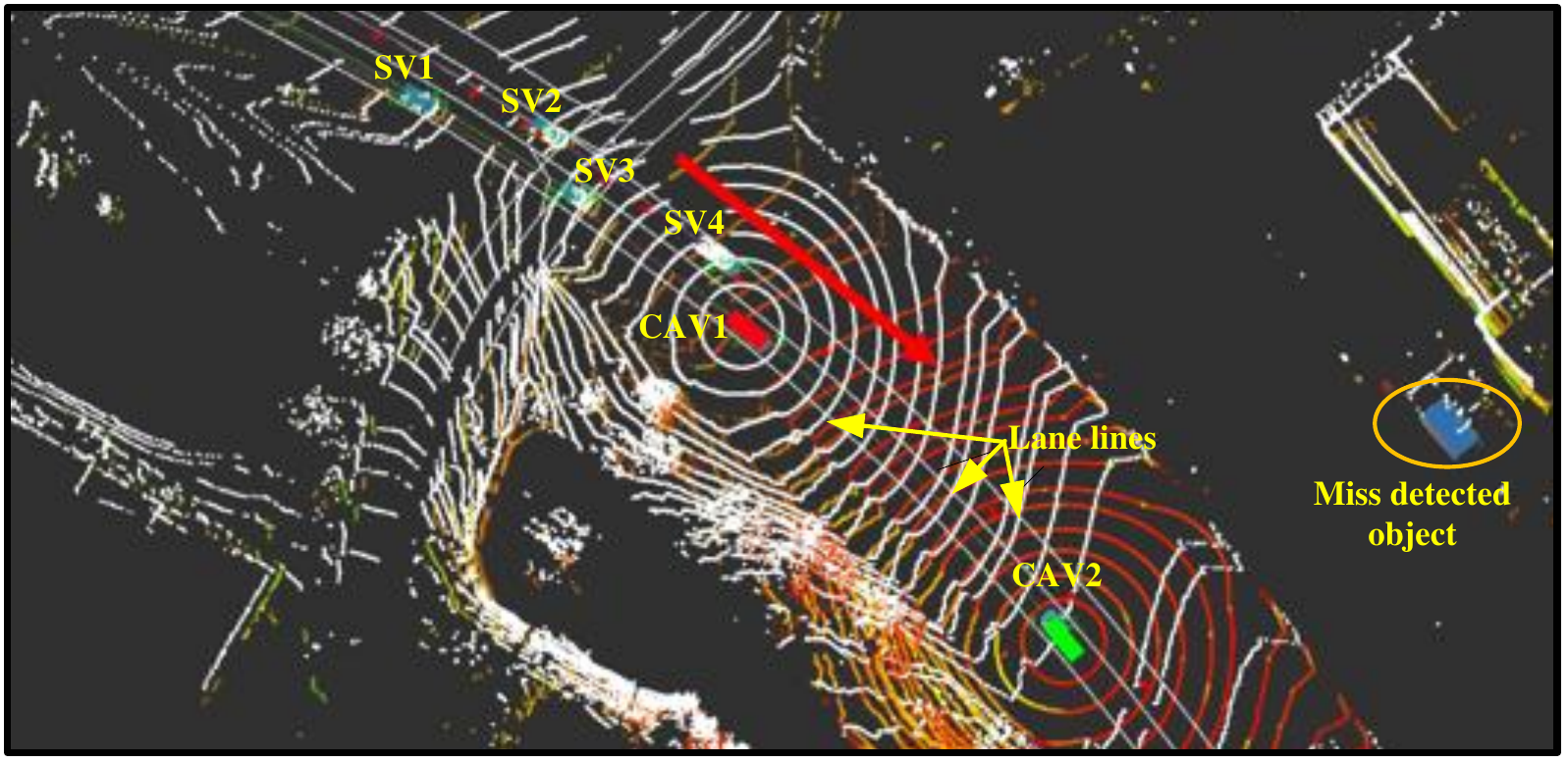}
\caption{Miss detected objects filtering.}
\label{fig: Miss detected objects filtering}
\end{figure}

\begin{figure}[htb!]
\centering
\includegraphics[width=0.8\linewidth]{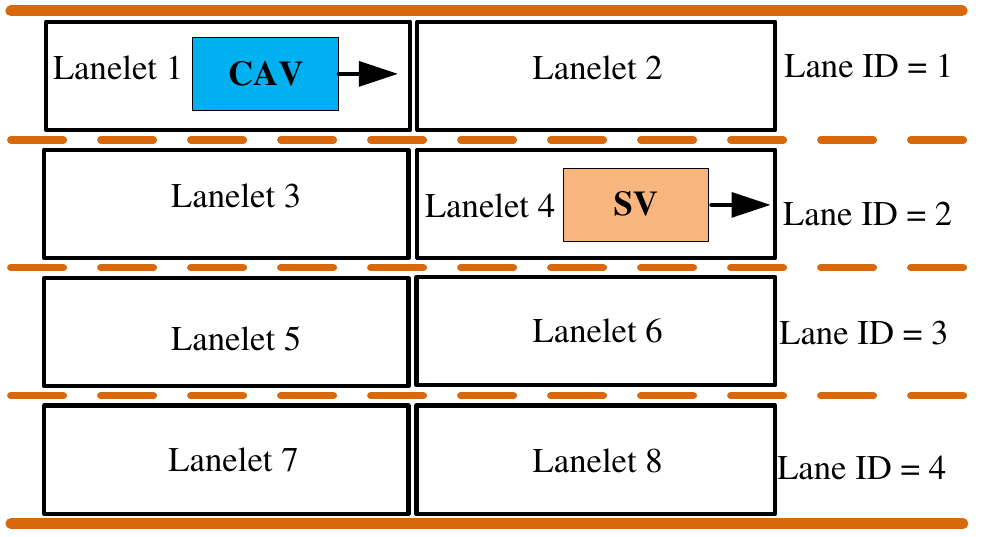}
\caption{Lanelet definition. CAV means connected automated vehicle, SV means subject vehicle. The lanelet ID and lane information for SV are 4 and 2, respectively. }
\label{fig: Lanelet definition}
\end{figure}

With the generated point cloud map, the vector map generation will be discussed in this subsection. The point cloud map is imported to RoadRunner and then, the road is drawn based on the exact lane position which can be inferred by the intensity information in the point cloud map as elements with a different color on the road surface have different intensity information which can be visualized in RoadRunner. With the lane lines drawn, the opendrive (Xodr) format map is output. In addition, a opendrive to lanelet map converter \cite{rehrl2022towards} is used to convert the opendrive map to lanelet map. The lanelet map is used for two purposes: 1) The lanelet map can provide the constraints to filter out the objects off the road for object detection and tracking modules as the off-road objects are not of interest because they have no interaction with the subject vehicles. As shown in Fig.~\ref{fig: Miss detected objects filtering}, the object highlighted in the yellow circle is miss detected and can be filtered out by the white lane lines in the vector map. 2) It will provide the lanelet information of the road and this information is applied to provide the coordinates of the objects in Frenet coordinate \cite{poggenhans2018lanelet2} \cite{vogl2020frenet}. Fig.~\ref{fig: Lanelet definition} shows an example of lanelet map where the lanes are segmented into lanelets and each lane has a unique ID.

\subsection{World model}\label{sec:world model}

The world model in CARMA platform \cite{soleimaniamiri2021cooperative} is applied to obtain the coordinates of the objects in Frenet coordinates. The world model is an interface that provides the supported access functions for working with the lanelet map and objects route, such as computing downtrack and crosstrack distances. The world model is tightly coupled to the lanelet2 library and relies on lanelet2 primitives for functionality. As shown in Fig.~\ref{fig: Frenet coordinates}, the downtrack distance is the $x$ in Frenet coordinates which is along the longitudinal direction of the road. The crosstrack distance is the $y$ in Frenet coordinate which is along the lateral direction of the road. The downtrack and crosstrack information of the objects can be used to analyze the car following behavior and lane change behavior of the objects with respect to the road coordinates.

\begin{figure}[htb!]
\centering
\includegraphics[width=0.5\linewidth]{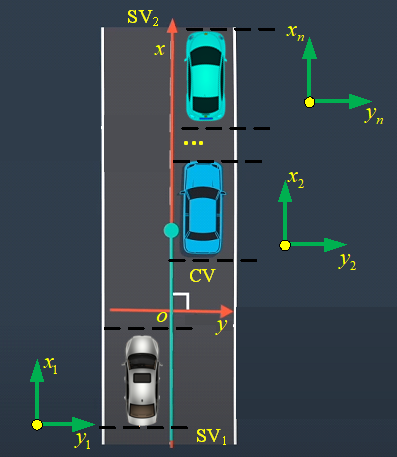}
\caption{Frenet coordinates.}
\label{fig: Frenet coordinates}
\end{figure}

As a downstream module of the object tracking algorithm, the world model will take the coordinates $x_m$ and $y_m$ of the on-road objects in the map coordinates and then, transform them into the coordinates in the Frenet coordinate. Accordingly, the lane and lanelet information where each object is will be obtained. Finally, these pieces of information will be output in the CSVs along with the information from the object tracking algorithm.

\section{Post processing}\label{sec:Post processing}

\begin{figure*}[htb!]
\centering
\includegraphics[width=1\linewidth]{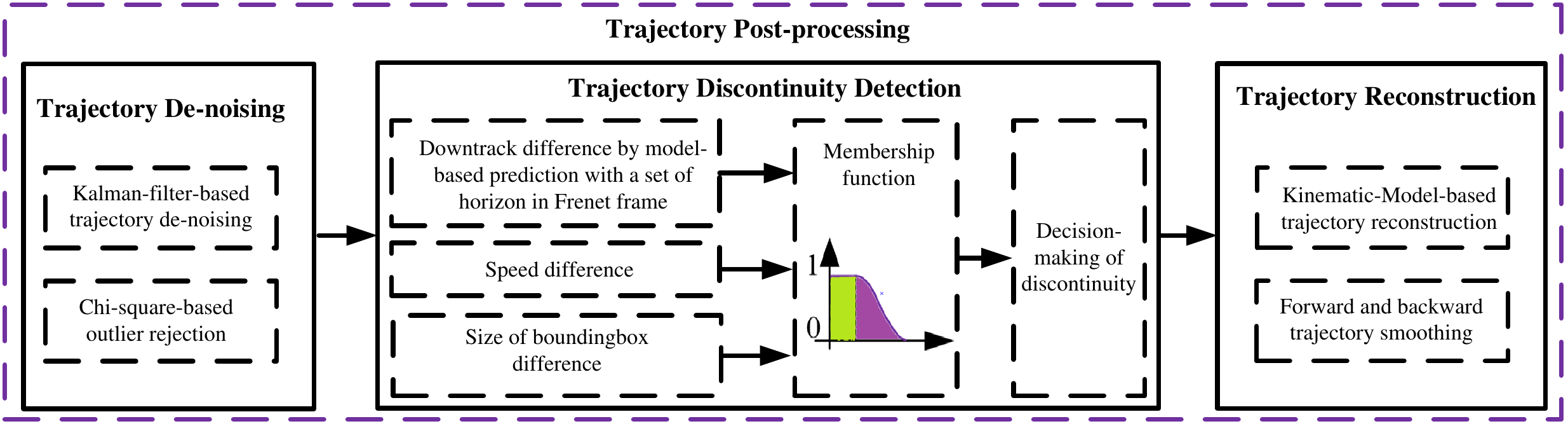}
\caption{Post-processing pipeline.}
\label{fig: Post-processing pipeline}
\end{figure*}

The trajectories from the object tracking algorithm may be noisy and contain outliers because the bounding boxes are not predicted accurately. On another aspect, the trajectories belonging to the same object might be disconnected due to the ID switch issue. To address these two issues, the post-processing algorithms are proposed to reduce the noise in the trajectories and connect the discontinuous trajectories. The post-processing procedures are shown in Fig.~\ref{fig: Post-processing pipeline}.

\subsection{Trajectory de-noising and outlier rejection}\label{sec:trajectory denoising and ourlie rejection}

There will be errors in the 3D bounding boxes due to the sparsity of LiDAR point cloud, which will lead to noise and outliers in the original trajectories from the object detection and tracking framework. To reduce the noise and correct the outliers, in this subsection, the denoising and outlier rejection methods are introduced based on the Kalman filter and Chi-square test algorithms \cite{wang2016chi} because of their good performance and ease of usage. The position in both map and Frenet coordinates will be processed through this post-process.

To implement the Kalman filter, state variables $\boldsymbol{x_m}=[x_m, y_m, \varphi_m, v_m, a_m]^\top$ in the map coordinates are chosen. $v$ and $a$ are the total speed and acceleration in the horizontal plane of the map coordinate. The output $\boldsymbol{y}_m=[x_m, y_m, \varphi_m, v_m]^\top$ from the object tracking algorithm can be measurements to implement the Kalman filter. The system dynamic model is the same as the one used in \cite{himmelsbach2012tracking}. Details about implementing a Kalman filter can be found in \cite{xia2021autonomous}. The covariance matrices of the process noise and measurement noise in the Kalman filter will be tuned to estimate the states $\boldsymbol{x}_m$ with smaller noise.

When applying the Kalman filter, the innovation vector defined by \eqref{eqn:innovation} can be used to detect the outlier in the measurement variables. The innovation vector $\boldsymbol{\zeta}$ reflects the difference between the fresh measurement and the prediction states of the Kalman filter. When there is no outlier, the noise of the innovation vector should satisfy the zero mean Gaussian distribution. When there are outliers in the measurements which are from the object tracking algorithm, the mean value of the innovation is no longer equal to zero. 
\begin{align}
  	\boldsymbol{		\zeta}_k=\boldsymbol{y}_k-\boldsymbol{H}\boldsymbol{\hat x}_{k|k-1},
    \label{eqn:innovation}
\end{align}
where $\boldsymbol{H}$ is the measurement matrix of the Kalman filter and $\boldsymbol{\hat x}_{k|k-1}$ is the prediction state in time $k$. In addition, a statistics parameter used for outlier detection \cite{rife2013effect} is 
\begin{align}
  	{\lambda}_k=\boldsymbol{\zeta}_k^\top \boldsymbol{A}_k^{-1}\boldsymbol{\zeta}_k,
    \label{eqn:statistics variable}
\end{align}
If the dimension of measurement $l$ and the statistics parameter ${\lambda}_k$ follow a Chi-square distribution with $l$ degrees of freedom, i.e., ${\lambda}_k \sim \chi^2(l)$, given a specific alarm rate $P_f$, then a threshold $T$ of the outlier detection can be obtained according to the Chi-square distribution. Then, the detection criteria can be \eqref{eqn:chi square condition}
\begin{align}
    \setlength{\arraycolsep}{4pt}
         \begin{cases} 
                        {\lambda}_k<T, \ Normal \\ 
                        {\lambda}_k \ge T , \ Fault
          \end{cases},
    \label{eqn:chi square condition}
\end{align}
when ${\lambda}_k$ is smaller than $T$, the measurement variable $\boldsymbol{y}_m$ is normal or $\boldsymbol{y}_m$ is an outlier. When the outlier is detected, the measurement variable $\boldsymbol{y}_m$ will be isolated and the states will be replaced by the prediction in the Kalman filter. Then, the trajectories in the original CSVs will be refined in terms of the noise level and outlier rejection. 

Besides processing the objects' states in the map coordinate, the states in the Frenet coordinate are processed as well because the downtrack information is very useful when calibrating the car-following model or analyzing the car-following behavior. In the Frenet coordinate, $\boldsymbol{x}_f=[x_f, v_f, a_f]^\top$ are considered as states. The measurements $\boldsymbol{y}_f=[x_f, v_f]^\top$ are the output from the world model. The vehicle-kinematic-model in our previous work \cite{xiasecure} is used as the system dynamic model to implement the Kalman filter. Other procedures are the same as how to process the states in the map coordinate.

\subsection{Trajectory discontinuity detection} \label{sec:Trajectory discontinuity detection}

\begin{figure}[htb!]
\centering
\includegraphics[width=0.9\linewidth]{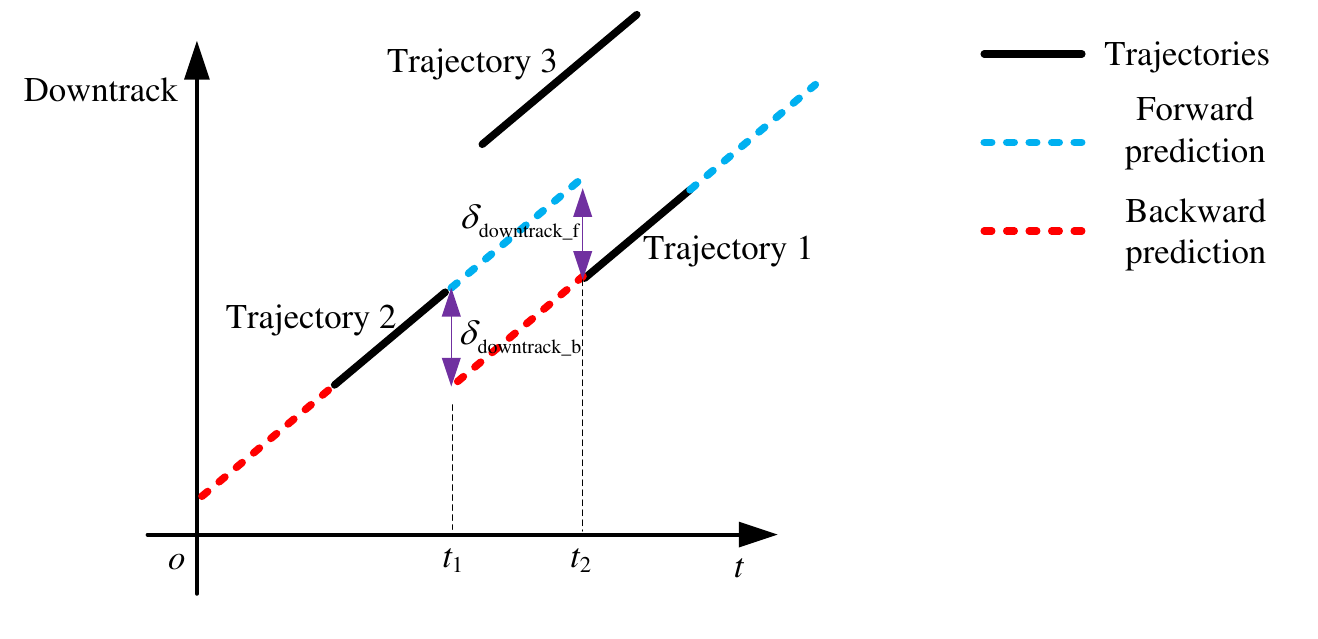}
\caption{Downtrack difference.}
\label{fig: Downtrack_difference.}
\end{figure}

As discussed in the object tracking section, an ID switch issue may happen, which leads to the discontinuity of the trajectories for the same object. In this subsection, a fuzzy-logic-based approach is proposed to detect the trajectories which belong to the same object. Given the objects' trajectories, the downtrack difference $\delta_{downtrack}$ between the predicted downtrack based on an existing trajectory and all other trajectories, speed difference $\delta_{speed}$ between the speed at the end point of the trajectory and the start point of the trajectory for two objects, and difference $\delta_{bb}$ between the bounding boxes of two objects will be input to the membership function \eqref{eqn:membership_function}.

\begin{align}
    \setlength{\arraycolsep}{4pt}
    \Theta_i = \begin{cases} 
                        1, & \text{abs}(\alpha) <\alpha_{\text{Thresh}}
                     \\ \frac{1}{\sigma_i\sqrt{2\pi}} e^{-\frac{(\text{abs}(\alpha)-\alpha_{\text{Thresh}})^2}{2\sigma_i^2}} ,& \text{abs}(\alpha) \ge \alpha_{\text{Thresh}}
          \end{cases},
    \label{eqn:membership_function}
\end{align}
where $\alpha$ is the input of the membership function and $\alpha_{\text{Thresh}}$ is a threshold.  Because the vehicle-kinematic model used for predicting downtrack might have discrepancies compared with the actual vehicle movement, $\delta_{downtrack}$ exists. If it is within a bound, the predicted trajectory and the existing trajectory may belong to the same object.  To obtain the downtrack difference $\delta_{downtrack}$, a vehicle-kinematic-model \cite{xiasecure} in \eqref{eqn:predicted downtrack} is used to predict the trajectory in a given time horizon, ie., to extend an existing trajectory in both forward and backward directions. For instance, in Fig.~\ref{fig: Downtrack_difference.}, for trajectory 2, a blue trajectory and a red trajectory will be predicted at the start (backward direction) and ending points (forward direction).
\begin{align}
    \setlength{\arraycolsep}{4pt}
        p=v \cdot \bigtriangleup t+p_{\text{0}},
    \label{eqn:predicted downtrack}
\end{align}
where $p$ is the predicted downtrack, $\bigtriangleup t$ ($\bigtriangleup t=\text{abs}(t_1-t_2)$ in Fig.~\ref{fig: Downtrack_difference.}) is the prediction time horizon, $p_0$ is the first or last point in the trajectory to be matched, and $v$ is the speed for the trajectory, which the prediction is based on. In our application, $\bigtriangleup t 	\in [-\tau, \tau]$, where $-\tau$ is the lower bound for the backward prediction and $\tau$ is the upper bound for the forward prediction. 

As $\delta_{downtrack}$ is an absolute distance which is related to the speed of the of vehicle, to normalize the downtrack difference information, \eqref{eqn:normalized downtrack} is used,
\begin{align}
    \setlength{\arraycolsep}{4pt}
        \alpha_{downtrack}=\text{abs}(\frac{\delta_{downtrack}}{v \cdot \bigtriangleup t   \cdot\eta}),
    \label{eqn:normalized downtrack}
\end{align}
where $\delta_{downtrack}=(\delta_{{downtrack}_\_b}+\delta_{{downtrack} _\_ f})/2$. $\eta$ is a speed error level used to normalize the downtrack. $\delta_{{downtrack} _\_ f}$ is the difference between the last point of the forward prediction and the first point of an existing trajectory (Trajectory 1 in Fig.~\ref{fig: Downtrack_difference.}). Accordingly, $\delta_{{downtrack}_\_b}$ is the difference between the first point of the backward prediction and the last point of an existing trajectory (Trajectory 2 in Fig.~\ref{fig: Downtrack_difference.}).  Taking the average of $\delta_{{downtrack}_\_b}$ and $\delta_{{downtrack} _\_ f}$ as $\delta_{downtrack}$ is a compromise to reduce error in the predicted trajectory. This is because, in \eqref{eqn:predicted downtrack}, a constant speed vehicle kinematic model is assumed to predict the trajectory. However, there might be speed variation during $\bigtriangleup t$ causing an error in the downtrack prediction.

The speed difference in \eqref{eqn:normalized velocity difference} between the two trajectories is another metric to detect if they belong to the same object or not.
\begin{align}
    \setlength{\arraycolsep}{4pt}
        \alpha_{speed}=1-\text{abs}(\frac{v-v_0}{ v}),
    \label{eqn:normalized velocity difference}
\end{align}
where $v$ is the speed of one trajectory and $v_0$ is the speed of the trajectory to be matched. The third index for the trajectory discontinuity detection is the difference defined in \eqref{eqn:normalized length difference} between the bounding boxes of the two objects having the two trajectories.
\begin{align}
    \setlength{\arraycolsep}{4pt}
        \alpha_{bb}=1-(\text{abs}(\frac{l-l_0}{l})+\text{abs}(\frac{w-w_0}{w}))/2,
    \label{eqn:normalized length difference}
\end{align}
where $l$ and $l_0$ are the length of the bounding boxes for two objects to be matched and $w$ and $w_0$ are the width of the bounding boxes for two objects to be matched. After having $\alpha_{downtrack}$, $\alpha_{speed}$, and $\alpha_{bb}$, they will be input to the membership function shown in \eqref{eqn:membership_function} to obtain $\Theta_{\text{downtrack}}$, $\Theta_{\text{speed}}$, and $\Theta_{\text{bb}}$. A convex combination motivated by \cite{cheli2007methodology} is applied to integrate corresponding weights for each index channel to obtain the  final index which is used to determine if the two trajectories belong to the same object. 
\begin{align}
    \setlength{\arraycolsep}{4pt}
    \Theta =k_1\Theta_{\text{downtrack}}+k_2\Theta_{\text{speed}}+k_3\Theta_{\text{bb}},
    \label{eqn:total_weights}
\end{align}
where $k_1$, $k_2$, and $k_3$ are the coefficients of the weights of $\Theta_{\text{downtrack}}$, $\Theta_{\text{speed}}$, and $\Theta_{\text{bb}}$, respectively. These parameters need to be tuned in real applications. If $\Theta$ is smaller than a defined threshold $\varrho$, the two trajectories belong to one object or they are for two objects. 

\begin{remark}
Note that all the trajectories will need to be searched to detect the discontinuity of trajectories. Another notice is that in \eqref{eqn:position_x}, the time horizon $\bigtriangleup t$ is a variable, which is unknown before the trajectory pair for the same object is found. Different time horizons of $\bigtriangleup t$ will be tried to search for the trajectory pair until the condition of the fuzzy logic condition is met. 
\end{remark}

\subsection{Trajectory reconstruction}
To fill the gap between two disconnected trajectories, a constant speed and steering vehicle kinematic model will be used to predict the vehicle trajectory within a short time horizon \cite{xiao2020vehicle}. For the old trajectory in the trajectory pair, forward prediction is performed according to \eqref{eqn:position_x} and \eqref{eqn:position_y}; for the new trajectory, backward prediction is performed. 
\begin{align}
    \setlength{\arraycolsep}{4pt}
        x_m=v_m\text{cos}(\varphi_m) \cdot \bigtriangleup t+x_{\text{0}_m},
    \label{eqn:position_x}
\end{align}
\begin{align}
    \setlength{\arraycolsep}{4pt}
        y_m=v_m\text{sin}(\varphi_m) \cdot \bigtriangleup t+y_{\text{0}_m},
    \label{eqn:position_y}
\end{align}
where $x_m$ and $y_m$ are in the map coordinate;$x_{{\text{0}}_m}$ and $y_{{\text{0}}_m}$ are the end and start point of the old and new trajectories, respectively; $\varphi_m$ is the heading angle of the object; $v_m$ is the speed of the object. $\bigtriangleup t$ is the prediction time horizon. As can be seen in \eqref{eqn:position_x} and \eqref{eqn:position_y}, the prediction errors in $x$ or $y$ directions increase with the length of the prediction time $\bigtriangleup t$ linearly. The errors come from the speed and heading error because the actual speed and heading might not be maintained as constants.  

\begin{figure}[htb!]
\centering
\includegraphics[width=0.8\linewidth]{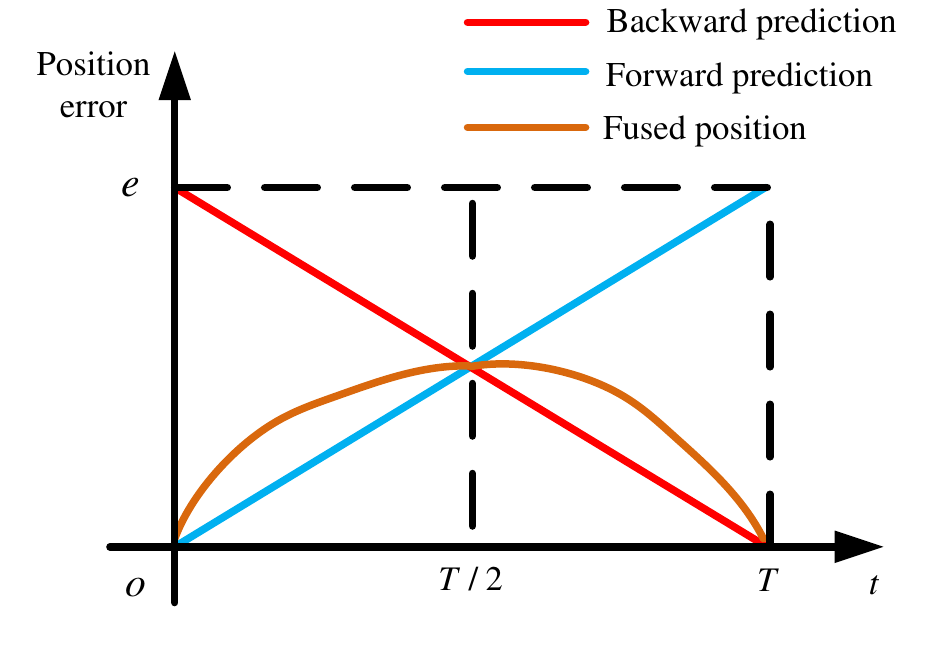}
\caption{Position error.}
\label{fig: Frenet coordinates}
\end{figure}
The position error propagation for the forward prediction and backward prediction is shown in Fig~\ref{fig: Frenet coordinates}. From time 0 to $T$, the prediction position error for the forward prediction increases, and from time $T$ to 0,  the prediction position error for the backward prediction decreases. The smaller the prediction time is, the smaller the prediction error is. Thus, to leverage the position from both forward and backward prediction,  \eqref{eqn:p_fused} is proposed to fuse the positions.
\begin{align}
    \setlength{\arraycolsep}{4pt}
        p_{\text{fused}}=\omega_{\text{forward}} \cdot p_{\text{forward}}+ \omega_{\text{backward}} \cdot p_{\text{backward}} ,
    \label{eqn:p_fused}
\end{align}
where $p$ denotes the $x_m$ or $y_m$. $\omega$ is the fusion weight of the position. The subscript "forward" denotes the position or weight is for the forward prediction. The subscript "backward" denotes the position or weight is for the backward prediction. The weights for the position of forward prediction and backward prediction are defined by \eqref{eqn:weight_forward} and \eqref{eqn:weight_backward}. 
\begin{align}
    \setlength{\arraycolsep}{4pt}
        \omega_{\text{forward}}=1-\frac{\bigtriangleup t}{T}, \bigtriangleup t 	\in [dt, T] ,
    \label{eqn:weight_forward}
\end{align}
where $T$ is the time interval between the two trajectories and $dt$ is the sample time of the trajectory which is 0.1s. 
\begin{align}
    \setlength{\arraycolsep}{4pt}
        \omega_{\text{backward}}=\frac{\bigtriangleup t}{T}, \bigtriangleup t 	\in [dt, T] ,
    \label{eqn:weight_backward}
\end{align}
It can be seen that at the beginning, i.e. $\bigtriangleup t$ is $dt$, and $\omega_{\text{forward}}$ is equal to one approximately, meaning that the fused position relies on the forward prediction. By using the fusion  The prediction position error is between the prediction position error and the backward position error. 

\setlength\belowdisplayskip{1pt}
\begin{table*}[h]
\renewcommand{\arraystretch}{1.3}
\caption{Output of data processing pipeline}
\label{tab:Output of data processing pipeline}
\centering
\begin{tabular}{c c c c}
\hline
\bfseries Variable & \bfseries Description & \bfseries Unit & \bfseries Coordinate \\
\hline
 ID    & Identity number for the surrounding vehicle &  N/A   & N/A   \\
\hline
 Time   & Timestamp (ascending by start time) of the corresponding row in CSV  & s & N/A \\
\hline
distance\_{sv} (headway) & Distance between the reference point of the reference CAV and the center point of the SV & m & N/A \\
\hline
pos\_x\_{sv}\_f    & x coordinate & m & Frenet \\
\hline
pos\_y\_{sv}\_f    & x coordinate & m & Frenet \\
\hline
pos\_x\_{sv}\_m    & x coordinate & m & map \\
\hline
pos\_y\_{sv}\_m    & y coordinate & m & map \\
\hline
heading\_{sv}\_m    & Azimuth  & deg & map \\
\hline
dim\_x\_{sv}   & length of the vehicle & m & vehicle \\
\hline
dim\_y\_{sv}   & width of the vehicle & m & vehicle \\
\hline
dim\_z\_{sv}   & height dimension of the surrounding vehicle & m & vehicle \\
\hline
speed\_{sv}   & Speed of the surrounding vehicle & m/s & map \\
\hline
acc\_{sv}   & Acceleration of the surrounding vehicle & $m/s^{2}$ & Frenet \\
\hline
closest\_{distance}\_{longitudinal}   & Relative longitudinal distance between the surrounding vehicle and the CAV vehicle & m & Frenet \\
\hline
closest\_distance\_lateral   & Relative lateral distance between the surrounding vehicle and the CAV vehicle & m & Frenet \\
\hline
lanelet\_id\_sv   & Lanelet ID of the surrounding vehicle’s center point & N/A & N/A \\
\hline
lane\_id\_sv   &Lane ID of the surrounding vehicle’s center point & N/A & N/A \\
\hline
total\_lanes   & Total lanes at the current position & N/A & N/A \\
\hline

\end{tabular}
\footnotesize{

N/A means not applicable. The reference CAV denotes the Tesla vehicle. }\\
\end{table*}

\section{Results and discussion}\label{sec:Results and discussion}
In this subsection, visualization results of the object detection and tracking pipeline and quantitative results of trajectory post-processing are discussed. 
\subsection{Experiment scenario and output of the data processing framework}
The two retrofitted CAVs were deployed on naturalistic scenarios for several hours with sensor data collected. Specifically, the collected data for this work include 2.93 hours Freeway scenario data, 2.07 hours city road scenario data, and 3.41 hours highway scenario data. Specifically, the freeway scenarios have merging onto freeway and exiting freeway cases, the city road and highway scenarios include passing intersections, lane change, left and right turn scenarios. All these data have been processed and the processed data will be published in the near future. The processed data are saved in CSV format which is similar to that for NGSIM datasets. Especially, the datasets are the complement of our precious simulated CAV datasets OPV2V \cite{xu2021opv2v}, which is the first published real dataset for CAV cooperative perception development. The output information from our data acquisition and processing platform is listed in Table.~\ref{tab:Output of data processing pipeline}. As can be seen in Table.~\ref{tab:Output of data processing pipeline}, a unique ID of each detected SV within the detection range of two CAVs will be output in the table. Along with the ID, the time of each LiDAR frame which has the corresponding SV is output. Then, information such as headway between the reference CAV which is the Tesla vehicle, and the SV, position, orientation (heading), speed and acceleration of the SV in the map frame,  coordinates in the Frenet frame, dimension of the SV, the longitudinal and lateral distance between the SV and CAV, and the lane and lanelet ID as well as the total lanes where the SV is are provided. These pieces of information are presented in the CSV format and are ease of usage for the transportation community in terms of CAV/AV and human-driven vehicle interaction analysis, car-following behavior analysis, traffic flow modeling as well as scenario generation purposes.

\subsection{Visualization of test scenario, HD map, and object detection and tracking
}
\begin{figure*}[htb!]
\centering
\includegraphics[width=1\linewidth]{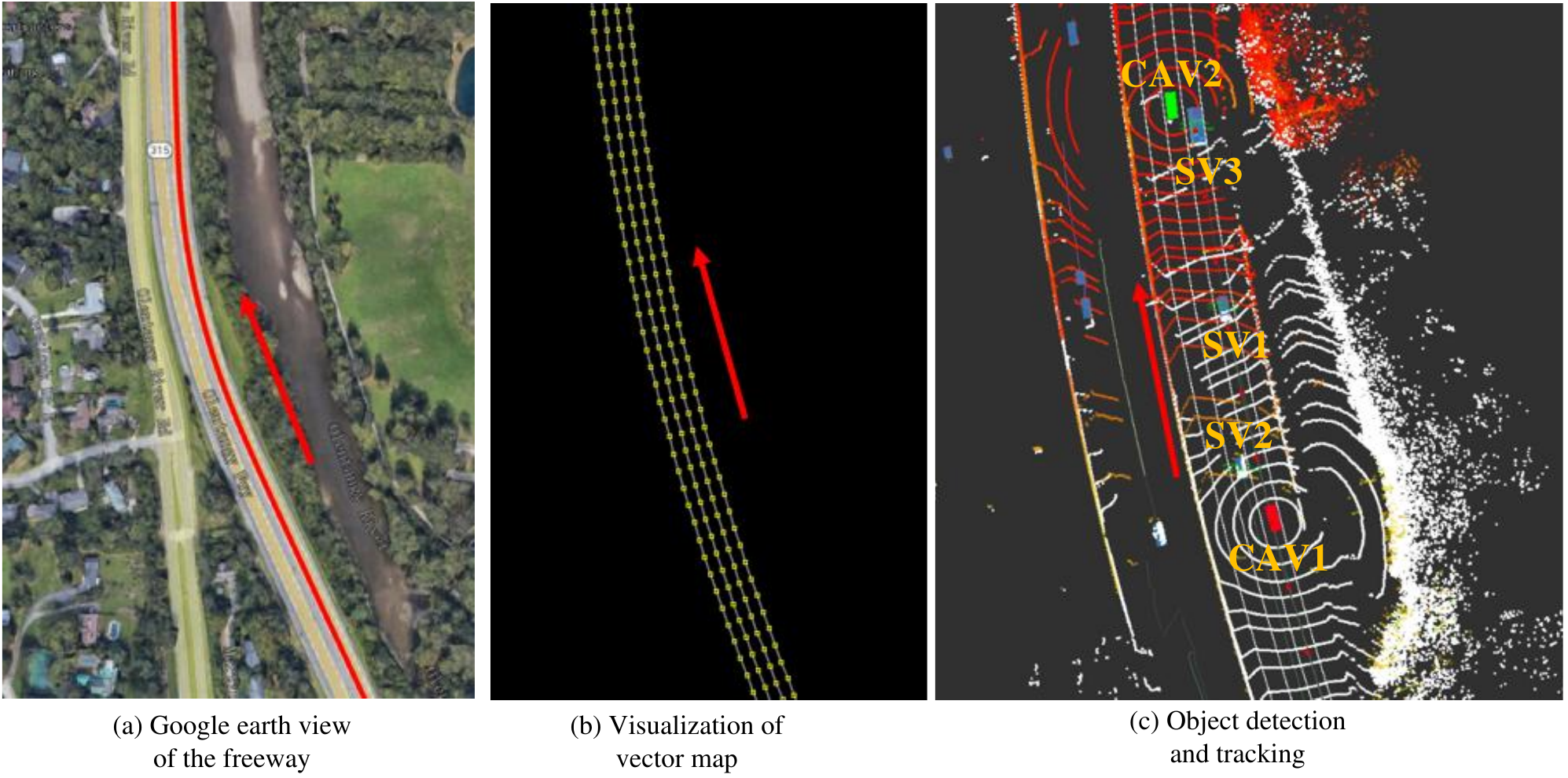}
\caption{Visualization of test scenario, vector map, and object detection and tracking. The red arrow in these three sub-figures shows the moving direction of the vehicle. In (c), the colored and white point clouds are from the LiDARs in the Tesla vehicle and AutonomouStuff vehicle, respectively. The blue bounding boxes are the object detection results of the surrounding vehicles. The red and green boxes are the AutonomouStuff and Tesla vehicles, respectively.
}
\label{fig:Visualization_object_detection_tracking}
\end{figure*}

Parts of the output from the data processing framework are shown in Fig.~\ref{fig:Visualization_object_detection_tracking}. In Fig.~\ref{fig:Visualization_object_detection_tracking}.a is the view of Google Earth to visualize the freeway scenario, the red line shows the trajectory of CAV1, and the operation speed is about 112km/h. Fig.~\ref{fig:Visualization_object_detection_tracking}.b visualizes the vector map. As can be seen from it, there are three lanes in this freeway scenario. Fig.~\ref{fig:Visualization_object_detection_tracking}.c provides the visualization of the object detection and tracking results. The colored point cloud is from the LiDAR in CAV1 and the white point cloud is from CAV2. The blue bounding boxes are the detection results of SVs. The green and red boxes represent the CAV1 and CAV2, respectively. In the lane where CAV2 is, SV2 is behind the vehicle which is out of the detection range of CAV2 but can be sensed by the LiDAR in CAV1. This is the benefit of the cooperative perception introduced in Section.~\ref{sec:Late fusion of object detection}. The same case happens to SV3 which is out of the detection range of CAV1 but detected by CAV2. In other words, through cooperative perception, the detection range of each CAV has been extended and the blind spot of a specific CAV can be covered by other CAVs. In the area where both CAVs cover, SV1 can be detected by both CAVs and the object detection results are fused and only one bounding box is output, which confirms the effectiveness of our late fusion strategy. Thanks to this cooperative perception, the sensing area of each CAV has been expanded and more SVs can be detected which means the output of the data processing framework provides the potential to analyze complex interactions between CAV and multiple SVs. For instance, in the lane where CAV1 is, CAV1, SV1, and SV3 can be used to investigate the car-following behavior of both CAV1 and SV1.

\begin{figure}[htb!]
\centering
\includegraphics[width=0.9\linewidth]{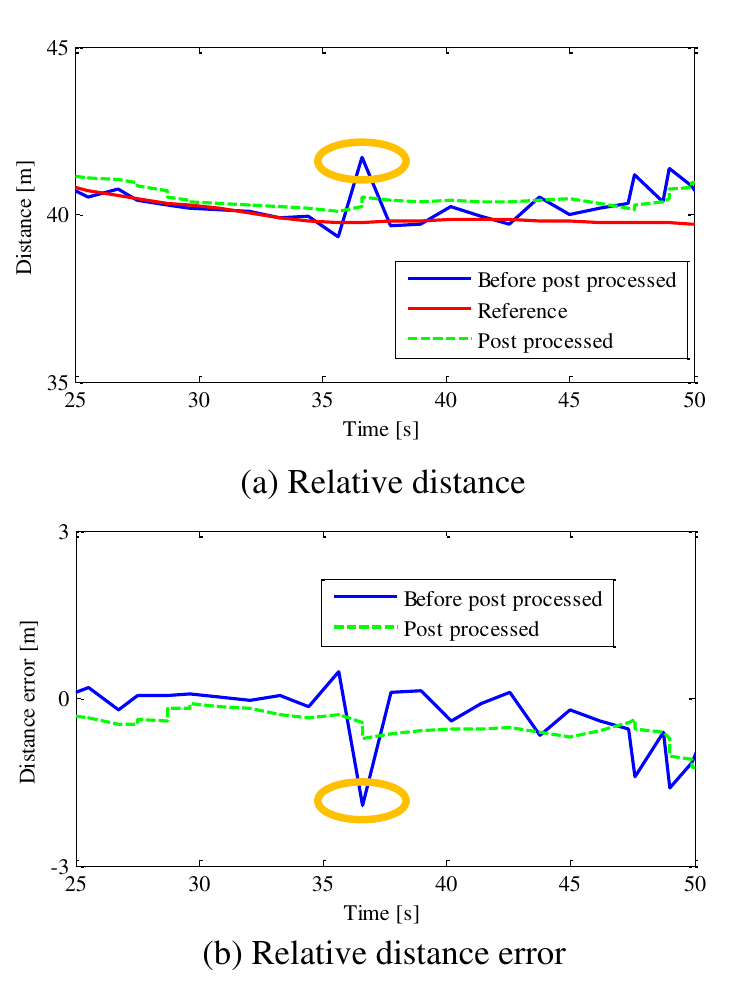}
\caption{Relative distance and its error. 
}
\label{fig:distance_results}
\end{figure}

\begin{figure}[htb!]
\centering
\includegraphics[width=1.1\linewidth]{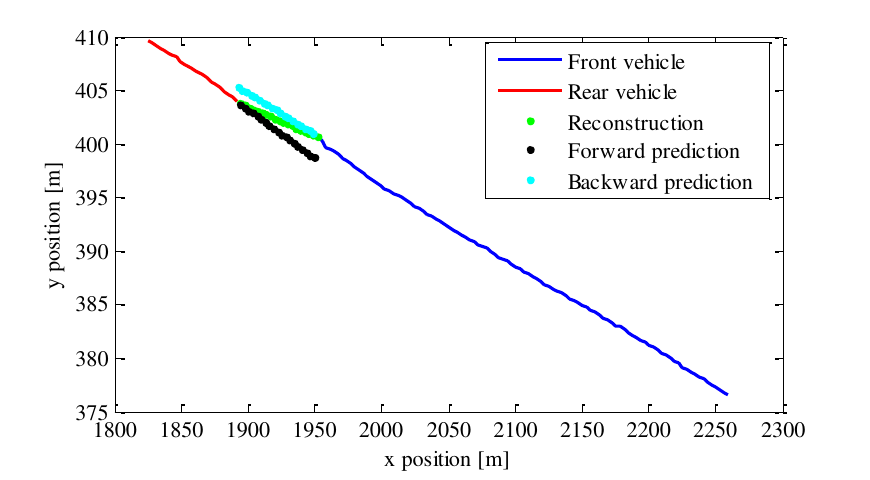}
\caption{Trajectory reconstruction. The blue and red lines represent the trajectories for the front vehicle and rear vehicle, respectively. The cyan and black dot lines are the backward and forward predicted trajectories respectively. The green dot line is the forward-backward-prediction fused trajectory, i.e., the reconstructed trajectory. 
}
\label{fig:Trajectory prediction}
\end{figure}

\subsection{Trajectory de-noising
}

Since the inter-distance information ($\text{distance}\_\text{sv}$ in Table.~\ref{tab:Output of data processing pipeline}) is of interest and importance to the transportation community, in this subsection, the results of the inter-distance accuracy are discussed. When calculating the inter-distance between the CAV and SV to analyze the car-following behavior, the error in the objects' position will be transferred to the inter-distance and accordingly affect the behavior analysis. In other words, the accuracy of the inter-distance also reflects the accuracy of the SV's trajectory. 
As each CAV is equipped with an RTK-corrected GNSS/IMU integration system, the position from this system can be used as the ground truth for the inter-distance. The inter-distance is calculated by obtaining the difference between the actual CAV position and the position of SV from the object tracking algorithm. We selected the case where one CAV is following the other CAV and is around 40m behind because in this range, the object detection module may predict objects with only a few point clouds, and the fitted bounding boxes may not be stable, leading to position error.

Fig.~\ref{fig:distance_results} shows the results the inter-distance and it error. In Fig.~\ref{fig:distance_results}.a, the ground truth is represented by the red line (Reference). The inter-distance computed by using the position directly from the object tracking algorithm is shown in the blue line (Before post processed). The green line (after processed) shows the inter-distance which is based on the post-processed position by the Kalman filter and Chi-square algorithm. Fig.~\ref{fig:distance_results}.b gives the inter-distance errors of Before post processed and Post processed. As can be seen, there are outliers highlighted by the yellow circle around $t=36$s in the blue line and the blue line is noisy. After post-processing, the green line is much smoother, meaning the noise is much smaller because not only the outlier is detected by the Chi-square algorithm but also the Kalman filter reduces the noise in the trajectory from the object tracking algorithm.
Similar results can be obtained from Fig.~\ref{fig:distance_results}.b and we can see that the maximum inter-distance error is smaller than 1m which is also smaller than that of the blue line. Therefore, the results in Fig.~\ref{fig:distance_results} proved the effectiveness of the trajectory de-noising algorithm in the post-processing module. The enhanced and smooth trajectory is more suitable for applications such as car-following behavior analysis.

\subsection{Trajectory Reconstruction
}
On top of the trajectory de-noising, the ID switch issue in the object tracking algorithm will cause a trajectory disconnection issue leading the trajectories belonging to the same object segmented. Our proposed trajectory discontinuity algorithm is able to find the disconnected trajectories and send the disconnected trajectories to the trajectory reconstruction module to recover the trajectory by using the kinematics-model-based forward and backward prediction algorithm. The results of the trajectory reconstruction in the map coordinate are provided in Fig.~\ref{fig:Trajectory prediction} in a freeway scenario. The red and blue trajectories are detected as disconnected and they belong to the same object. Then, the backward prediction shown by the cyan dot line in Fig.~\ref{fig:Trajectory prediction} is made for the front vehicle, and the forward prediction represented by the black dot line in Fig.~\ref{fig:Trajectory prediction} is made for the rear vehicle. Since the prediction is based on the kinematic-vehicle-model, during the prediction, the velocity and heading are assumed as constants. However, in real application, they are not usually rigorously constants. The difference between the actual speed and heading will contribute to the prediction error, i.e., the difference between the actual trajectory and the predicted trajectory. As illustrated by Fig.~\ref{fig:Trajectory prediction}, there is a difference between the end of the black line and the start of the blue line. The proposed forward-backward-prediction fusion method takes advantage of the trajectories of both cyan and black lines when they have good accuracy and gives the fused trajectory result shown by the green dot line in Fig.~\ref{fig:Trajectory prediction}. It is obvious that the green dot line connects the blue and red lines more smoothly than both the cyan and black dot lines.

\begin{remark}
It should be noted that although the results in this section have proven the functionality of the proposed ADS sensor data acquisition and processing platform, there is still space for improvement. For example, in the near future, LiDARs with denser lasers can be tried to improve object detection performance, and also, the deep-learning-based approach can be applied to involve the intention information in the trajectory reconstruction module to improve the trajectory.
\end{remark}

\section{Conclusions}\label{sec:conclusions}

In this paper, an ADS data processing framework for vehicle trajectory extraction, reconstruction, and evaluation is proposed. The following conclusions can be made: 1) The proposed framework is capable of processing the advanced sensor data from multi-CAVs, i.e., detecting the objects around CAVs and tracking the objects to extract the objects’ ID, position, speed, and orientation information in different coordinates. 2) The world model with HD maps can assist the object detection and tracking algorithm in filtering out the off-road objects and obtaining the downtrack and crosstrack information in Frenet coordinate; 3) The noise and outlier in the trajectories can be reduced and rejected by the Kalman filter and the Chi-square test method to further improve the objects' information such as trajectories; the proposed trajectory discontinuity detection method can identify the discontinuous trajectories and the trajectories can be reconstructed by our forward-backward-prediction smoothing method.

Future work can focus on involving camera data in the data processing framework to improve object detection robustness, adding infrastructure data to enrich the capability of the data processing framework in intersection scenarios, and developing deep learning algorithms to further enhance multi-modal multi-vehicle sensor fusion to improve the performance.

\appendices
\section{Map coordinate definition}  \label{sec:map coordinate definition}

\begin{figure}[htb!]
\centering
\includegraphics[width=0.8\linewidth]{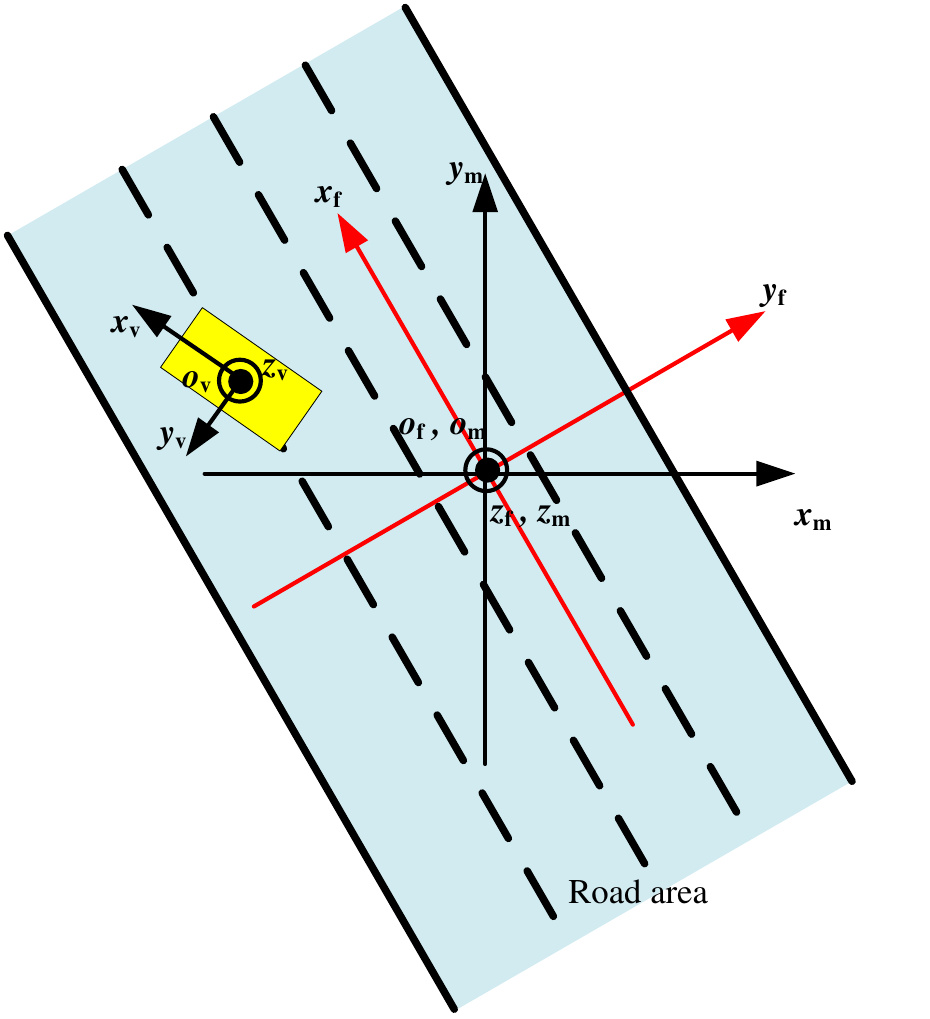}
\caption{Coordinates definition.}
\label{fig: Coordinates definition}
\end{figure}

The map frame is a fixed frame attached to a certain selected point and $x_m$, $y_m$, and $z_m$ are towards east, north, and upward directions, respectively, as shown in Fig.~\ref{fig: Coordinates definition}. Note that, we usually pick the start point of a CAV’s route as the origin of the map frame. The unit of the coordinate in this frame is meter.

\section{Frenet coordinate definition}  \label{sec:frenet coordinate definition}

The Frenet frame is attached to the center of the road and its orientation will change as the road direction changes. And the $x_f$ and $y_f$ mean the longitudinal forward and lateral right directions of the road as shown in Fig.~\ref{fig: Coordinates definition}. The unit of the coordinate in this frame is the meter.

\section{Vehicle coordinate definition}  \label{sec:vehicle coordinate definition}

The vehicle frame is attached to the center of the rear axle of the CAV and will move and rotate as the CAV. As can be seen from Fig.~\ref{fig: Coordinates definition}, the $x_v$, $y_v$, and $z_v$ of the vehicle frame are towards the front, left, and upward directions. The unit of the coordinate in this frame is the meter.




\bibliography{Refs}

\begin{thebibliography}{10}

\bibitem{xu2022v2x}
R.~Xu, H.~Xiang, Z.~Tu, X.~Xia, M.-H. Yang, and J.~Ma, ``V2x-vit:
  Vehicle-to-everything cooperative perception with vision transformer,'' {\em
  arXiv preprint arXiv:2203.10638}, 2022.

\bibitem{xu2022cobevt}
R.~Xu, Z.~Tu, H.~Xiang, W.~Shao, B.~Zhou, and J.~Ma, ``Cobevt: Cooperative
  bird's eye view semantic segmentation with sparse transformers,'' {\em arXiv
  preprint arXiv:2207.02202}, 2022.

\bibitem{li2021operational}
T.~Li, X.~Han, J.~Ma, M.~Ramos, and C.~Lee, ``Operational safety of automated
  and human driving in mixed traffic environments: A perspective of
  car-following behavior,'' {\em Proceedings of the Institution of Mechanical
  Engineers, Part O: Journal of Risk and Reliability}, p.~1748006X211050696,
  2021.

\bibitem{hu2022processing}
X.~Hu, Z.~Zheng, D.~Chen, X.~Zhang, and J.~Sun, ``Processing, assessing, and
  enhancing the waymo autonomous vehicle open dataset for driving behavior
  research,'' {\em Transportation Research Part C: Emerging Technologies},
  vol.~134, p.~103490, 2022.

\bibitem{han2022strategic}
X.~Han, R.~Xu, X.~Xia, A.~Sathyan, Y.~Guo, P.~Bujanovi{\'c}, E.~Leslie,
  M.~Goli, and J.~Ma, ``Strategic and tactical decision-making for cooperative
  vehicle platooning with organized behavior on multi-lane highways,'' {\em
  Transportation Research Part C: Emerging Technologies}, vol.~145, p.~103952,
  2022.

\bibitem{dot2016next}
U.~DOT, ``Next generation simulation (ngsim) vehicle trajectories and
  supporting data,'' 2016.

\bibitem{jiang2014traffic}
R.~Jiang, M.-B. Hu, H.~Zhang, Z.-Y. Gao, B.~Jia, Q.-S. Wu, B.~Wang, and
  M.~Yang, ``Traffic experiment reveals the nature of car-following,'' {\em
  PloS one}, vol.~9, no.~4, p.~e94351, 2014.

\bibitem{wu2019tracking}
F.~Wu, R.~E. Stern, S.~Cui, M.~L. Delle~Monache, R.~Bhadani, M.~Bunting,
  M.~Churchill, N.~Hamilton, B.~Piccoli, B.~Seibold, {\em et~al.}, ``Tracking
  vehicle trajectories and fuel rates in phantom traffic jams: Methodology and
  data,'' {\em Transportation Research Part C: Emerging Technologies}, vol.~99,
  pp.~82--109, 2019.

\bibitem{zhaoopen}
D.~Zhao, X.~Li, X.~Shi, H.~Yao, R.~James, D.~K. Hale, and A.~Ghiasi, ``An open
  database generation with monte carlo based lane marker detection and critical
  analysis of vehicle trajectory-high-granularity highway simulation
  (high-sim),''

\bibitem{geiger2013vision}
A.~Geiger, P.~Lenz, C.~Stiller, and R.~Urtasun, ``Vision meets robotics: The
  kitti dataset,'' {\em The International Journal of Robotics Research},
  vol.~32, no.~11, pp.~1231--1237, 2013.

\bibitem{sun2020scalability}
P.~Sun, H.~Kretzschmar, X.~Dotiwalla, A.~Chouard, V.~Patnaik, P.~Tsui, J.~Guo,
  Y.~Zhou, Y.~Chai, B.~Caine, {\em et~al.}, ``Scalability in perception for
  autonomous driving: Waymo open dataset,'' in {\em Proceedings of the IEEE/CVF
  conference on computer vision and pattern recognition}, pp.~2446--2454, 2020.

\bibitem{houston2020one}
J.~Houston, G.~Zuidhof, L.~Bergamini, Y.~Ye, L.~Chen, A.~Jain, S.~Omari,
  V.~Iglovikov, and P.~Ondruska, ``One thousand and one hours: Self-driving
  motion prediction dataset,'' {\em arXiv preprint arXiv:2006.14480}, 2020.

\bibitem{caesar2020nuscenes}
H.~Caesar, V.~Bankiti, A.~H. Lang, S.~Vora, V.~E. Liong, Q.~Xu, A.~Krishnan,
  Y.~Pan, G.~Baldan, and O.~Beijbom, ``nuscenes: A multimodal dataset for
  autonomous driving,'' in {\em Proceedings of the IEEE/CVF conference on
  computer vision and pattern recognition}, pp.~11621--11631, 2020.

\bibitem{pitropov2021canadian}
M.~Pitropov, D.~E. Garcia, J.~Rebello, M.~Smart, C.~Wang, K.~Czarnecki, and
  S.~Waslander, ``Canadian adverse driving conditions dataset,'' {\em The
  International Journal of Robotics Research}, vol.~40, no.~4-5, pp.~681--690,
  2021.

\bibitem{huang2018apolloscape}
X.~Huang, X.~Cheng, Q.~Geng, B.~Cao, D.~Zhou, P.~Wang, Y.~Lin, and R.~Yang,
  ``The apolloscape dataset for autonomous driving,'' in {\em Proceedings of
  the IEEE conference on computer vision and pattern recognition workshops},
  pp.~954--960, 2018.

\bibitem{xu2022opv2v}
R.~Xu, H.~Xiang, X.~Xia, X.~Han, J.~Li, and J.~Ma, ``Opv2v: An open benchmark
  dataset and fusion pipeline for perception with vehicle-to-vehicle
  communication,'' in {\em 2022 International Conference on Robotics and
  Automation (ICRA)}, pp.~2583--2589, IEEE, 2022.

\bibitem{sharma2021connected}
A.~Sharma and Z.~Zheng, ``Connected and automated vehicles: Opportunities and
  challenges for transportation systems, smart cities, and societies,'' {\em
  Automating Cities}, pp.~273--296, 2021.

\bibitem{quigley2009ros}
M.~Quigley, K.~Conley, B.~Gerkey, J.~Faust, T.~Foote, J.~Leibs, R.~Wheeler,
  A.~Y. Ng, {\em et~al.}, ``Ros: an open-source robot operating system,'' in
  {\em ICRA workshop on open source software}, vol.~3, p.~5, Kobe, Japan, 2009.

\bibitem{luo2021exploring}
C.~Luo, X.~Yang, and A.~Yuille, ``Exploring simple 3d multi-object tracking for
  autonomous driving,'' in {\em Proceedings of the IEEE/CVF International
  Conference on Computer Vision}, pp.~10488--10497, 2021.

\bibitem{naujoks2018orientation}
B.~Naujoks and H.-J. Wuensche, ``An orientation corrected bounding box fit
  based on the convex hull under real time constraints,'' in {\em 2018 IEEE
  Intelligent Vehicles Symposium (IV)}, pp.~1--6, IEEE, 2018.

\bibitem{apollo}
``{Apollo}: an open autonomous driving platform.''
  https://github.com/ApolloAuto/apollo, 2017.

\bibitem{himmelsbach2012tracking}
M.~Himmelsbach and H.-J. Wuensche, ``Tracking and classification of arbitrary
  objects with bottom-up/top-down detection,'' in {\em Intelligent Vehicles
  Symposium (IV), 2012 IEEE}, pp.~577--582, IEEE, 2012.

\bibitem{feng2019multi}
W.~Feng, Z.~Hu, W.~Wu, J.~Yan, and W.~Ouyang, ``Multi-object tracking with
  multiple cues and switcher-aware classification,'' {\em arXiv preprint
  arXiv:1901.06129}, 2019.

\bibitem{tian2019online}
W.~Tian, M.~Lauer, and L.~Chen, ``Online multi-object tracking using joint
  domain information in traffic scenarios,'' {\em IEEE Transactions on
  Intelligent Transportation Systems}, vol.~21, no.~1, pp.~374--384, 2019.

\bibitem{xia2022estimation}
X.~Xia, L.~Xiong, Y.~Huang, Y.~Lu, L.~Gao, N.~Xu, and Z.~Yu, ``Estimation on
  imu yaw misalignment by fusing information of automotive onboard sensors,''
  {\em Mechanical Systems and Signal Processing}, vol.~162, p.~107993, 2022.

\bibitem{rehrl2022towards}
K.~Rehrl, S.~Groechenig, T.~Piribauer, R.~Spielhofer, and P.~Weissensteiner,
  ``Towards a standardized workflow for creating high-definition maps for
  highly automated shuttles,'' {\em Journal of Location Based Services},
  pp.~1--33, 2022.

\bibitem{poggenhans2018lanelet2}
F.~Poggenhans, J.-H. Pauls, J.~Janosovits, S.~Orf, M.~Naumann, F.~Kuhnt, and
  M.~Mayr, ``Lanelet2: A high-definition map framework for the future of
  automated driving,'' in {\em 2018 21st international conference on
  intelligent transportation systems (ITSC)}, pp.~1672--1679, IEEE, 2018.

\bibitem{vogl2020frenet}
C.~Vogl, M.~Sackmann, L.~K{\"u}rzinger, and U.~Hofmann, ``Frenet coordinate
  based driving maneuver prediction at roundabouts using lstm networks,'' in
  {\em Computer Science in Cars Symposium}, pp.~1--9, 2020.

\bibitem{soleimaniamiri2021cooperative}
S.~Soleimaniamiri, X.~S. Li, H.~Yao, A.~Ghiasi, G.~Vadakpat, P.~Bujanovic,
  T.~Lochrane, J.~Stark, D.~Hale, S.~Racha, {\em et~al.}, ``Cooperative
  automation research: Carma proof-of-concept transportation system management
  and operations use case 4-dynamic lane assignment,'' tech. rep., United
  States. Federal Highway Administration, 2021.

\bibitem{wang2016chi}
R.~Wang, Z.~Xiong, J.~Liu, J.~Xu, and L.~Shi, ``Chi-square and sprt combined
  fault detection for multisensor navigation,'' {\em IEEE Transactions on
  Aerospace and Electronic Systems}, vol.~52, no.~3, pp.~1352--1365, 2016.

\bibitem{xia2021autonomous}
X.~Xia, E.~Hashemi, L.~Xiong, A.~Khajepour, and N.~Xu, ``Autonomous vehicles
  sideslip angle estimation: Single antenna gnss/imu fusion with observability
  analysis,'' {\em IEEE Internet of Things Journal}, 2021.

\bibitem{rife2013effect}
J.~H. Rife, ``The effect of uncertain covariance on a chi-square integrity
  monitor,'' {\em Navigation: Journal of The Institute of Navigation}, vol.~60,
  no.~4, pp.~291--303, 2013.

\bibitem{xiasecure}
X.~Xia, R.~Xu, J.~Ma, and Y.~Li, ``Secure cooperative localization for
  connected automated vehicles based on consensus estimation,''

\bibitem{cheli2007methodology}
F.~Cheli, E.~Sabbioni, M.~Pesce, and S.~Melzi, ``A methodology for vehicle
  sideslip angle identification: comparison with experimental data,'' {\em
  Vehicle System Dynamics}, vol.~45, no.~6, pp.~549--563, 2007.

\bibitem{xiao2020vehicle}
W.~Xiao, L.~Zhang, and D.~Meng, ``Vehicle trajectory prediction based on motion
  model and maneuver model fusion with interactive multiple models,'' {\em SAE
  International Journal of Advances and Current Practices in Mobility}, vol.~2,
  no.~2020-01-0112, pp.~3060--3071, 2020.

\bibitem{xu2021opv2v}
R.~Xu, H.~Xiang, X.~Xia, X.~Han, J.~Liu, and J.~Ma, ``Opv2v: An open benchmark
  dataset and fusion pipeline for perception with vehicle-to-vehicle
  communication,'' {\em arXiv preprint arXiv:2109.07644}, 2021.

\end{thebibliography}
\bibliographystyle{ieeetr}

\ifCLASSOPTIONcaptionsoff
  \newpage
\fi






\end{document}